# A Symmetric Keyring Encryption Scheme for Biometric Cryptosystems

Yen-Lung Lai, Jung-Yeon Hwang, Zhe Jin, Soohyong Kim, Sangrae Cho and Andrew Beng Jin Teoh

*Abstract*—**In this paper, we propose a novel biometric cryptosystem for vectorial biometrics named symmetric keyring encryption (SKE) inspired by Rivest's keyring model (2016). Unlike conventional biometric secret-binding primitives, such as fuzzy commitment and fuzzy vault, the proposed scheme reframes the biometric secret-binding problem as a fuzzy symmetric encryption problem with a notion called resilient vector pair. In this study, the pair resembles the encryption–decryption key pair in symmetric key cryptosystems. This notion is realized using the index of maximum hashed vectors—a special instance of the ranking-based locality-sensitive hashing function. With a simple filtering mechanism and $[m, k]$ Shamir's secret-sharing scheme, we show that SKE, both in theoretical and empirical evaluation, can retrieve the exact secret with overwhelming probability for a genuine input yet negligible probability for an imposter input. Though SKE can be applied to any vectorial biometrics, we adopt the fingerprint vector as a case of study in this work. The experiments have been performed under several subsets of FVC 2002, 2004, and 2006 datasets. We formalize and analyze the threat model of SKE that encloses several major security attacks.**

## I. Introduction

Conventional security systems rely on a deterministic model to ensure information security [1]. This model works efficiently under the presumption that a legitimate user always holds the consistency by presenting a constant, uniquely and specifiable cryptographic key. However, in practice, situations arise in which human and other factors undermine the possibility of exactness and uniqueness in a security cryptosystem. A deterministic security system may not work when fuzziness is involved, for example in biometric systems in which authentication takes place by an individual presenting a biological trait e.g. fingerprint. Fingerprint features, however, are prone to distortion due to the presence of noise during the acquisition process. This prompts the growth of *error-tolerant* cryptographic systems, which are considered to be more feasible under the circumstances in which deterministic models do not apply.

The integration of cryptographic schemes and biometrics, a special instance of error-tolerant cryptographic systems, is indeed not new. This line of study is called biometric cryptosystems or biometric encryption [2]. Biometric cryptosystems are devised to protect secrets such as private keys by either binding/retrieving secrets with biometrics or generating secrets from biometrics directly. Fuzzy commitment [3], fuzzy vault [4], and fuzzy extractor [5] are three primitive instances of biometric cryptosystems.

This work was supported by the Institute for Information & communications Technology Promotion (IITP) grant funded by the Korea government (MSIT) (No.2016-0-00097, Development of Biometrics-based Key Infrastructure Technology for On-line Identification).

Lai Yen Lung and Jin Zhe are with the School of Information Technology, Monash University Malaysia ({jin.zhe, yenlung.lai}@monash.edu).

J. Y. Hwang, S. Kim and S.R Cho are with Electronics and Telecommunications Research Institute (ETRI) ({videmot, lifewsky, sangrae}@etri.re.kr).

Andrew B. J. Teoh is with the School of Electrical and Electronic Engineering, College of Engineering, Yonsei University, Seoul, South Korea. (Corresponding author, e-mail: bjteoh@yonsei.ac.kr).

In this paper, we introduce a novel biometric cryptosystem for vectorial biometrics (biometric representation that appears in a fixed-size vector form) and secret binding called symmetric keyring encryption (SKE). SKE is an *error-tolerant symmetric encryption construct* that is motivated by the recent Rivest's keyring model [6]. The key ingredient of Rivest's keyring model is a notion coined resilient vector (RV) pair that is analogous to the random key pair in conventional symmetric cryptosystems. We illustrate the idea as follows. Let Alice encrypt a secret by a RV $\Omega(\mathbf{x})$, which is derived from a sample set $\mathbf{x}$, known as *keyring* [6]. An instance of keyring is biometrics. The model allows either Alice herself or another user Bob to decrypt the secret via $\Omega(\mathbf{x}')$ that is similar to $\Omega(\mathbf{x})$. Decryption succeeds with probability $\mathbb{P}(\Omega(\mathbf{x}) = \Omega(\mathbf{x}')) = S(\mathbf{x}, \mathbf{x}') > \tau$, where $S(\cdot, \cdot)$ is a similarity measure between $\mathbf{x}$ and $\mathbf{x}'$ and $\tau$ is a preset threshold value. On the contrary, anyone who attempts to decrypt the secret with keyring that is distinct from $\mathbf{x}$ would fail certainly if $S(\mathbf{x}, \mathbf{x}') < \tau$.

The paper is organized as follows. In Section II, we present two preliminaries relevant to SKE: index of maximum (IoM) hashing and Rivest's keyring model. In Section III, SKE is described in terms of binding and retrieving phases, followed by Section IV that elaborates the resilient property of SKE. The threat model and the associated security analyses are presented in Section V. In Section VI, we provide a thorough empirical evaluation and analysis on SKE based on the fingerprint vectors generated from FVC (Fingerprint Verification Competition) benchmark datasets. We also provide a comparison with other competing biometric cryptosystems. Lastly, concluding remarks are given in Section VII.

### A. Related Works

In this section, we review several biometric cryptosystem primitives relevant to our scheme. One of the notable works is by Juels and Wattenberg that put forward the concept of fuzzy commitment [3]. To be specific, consider a person with his biometric input of different but close readings $\mathbf{x} \in \mathcal{F}^n$ and $\mathbf{x}' \in \mathcal{F}^n$ in a finite field $\mathcal{F}$. For fuzzy commitment, binary biometric features i.e. $\mathcal{F} = 2$ is used to generate an offset $\boldsymbol{\delta} \in \mathcal{F}^n$, where $\boldsymbol{\delta} = \mathbf{x}' \oplus \mathbf{x}$. In this sense, the Hamming weight of the offset $||\boldsymbol{\delta}||$ indicates the Hamming distance of $\mathbf{x}$, and $\mathbf{x}'$. During the enrollment, given a random codeword $\mathbf{c} \in \mathcal{F}^k$, which is an encoded secret of size $\mathcal{F}^k$, one can conceal $\mathbf{c}$ by generating a secure sketch $\mathbf{ss} = \mathbf{c} \oplus \mathbf{x}$ with $\mathbf{x}$. The codeword $\mathbf{c}$ is then hashed with a one-way hash function $H$ and stored along with $\mathbf{ss}$ as a commitment $(H(\mathbf{c}), \mathbf{ss})$. In the verification stage, given $(H(\mathbf{c}), \mathbf{ss})$, we can compute the corrupted codeword such as $\mathbf{c}' = \mathbf{ss} \oplus \mathbf{x}' = \boldsymbol{\delta} \oplus \mathbf{c}$ with $\mathbf{x}'$. With a decoding algorithm $\text{Decode}_t$ with error tolerance capacity $t$, if $||\boldsymbol{\delta}|| \leq t$, $\mathbf{c}$ can be recovered from $\mathbf{c}'$ i.e. $\text{Decode}_t(\mathbf{c}') = \mathbf{c}$ . Finally, the decommitment is deemed successful if $H(\mathbf{c}) = H(\text{Decode}_t(\mathbf{c}'))$ . The fuzzy commitment scheme is conceptually simple and its error tolerance mechanism solely relies on error correction codes (ECCs) such as Bose–Chaudhuri–Hocquenghem (BCH), Reed–Solomon codes etc. [3]. The security of the fuzzy commitment scheme is attributed to the reconstruction complexity of codeword $\mathbf{c}$, provided an adversary has no knowledge of $\mathbf{c}$ or $\mathbf{x}$ from $H(\mathbf{c})$ under a random oracle model (ROM). A practical shortcoming of fuzzy commitment is the difficulty to provide a rigorous proof about security over the nonuniformity of $\mathbf{x}$.

Another major biometric cryptosystem primitive that also carries an error tolerance property is the fuzzy vault [4]. Let $k \leq n$. Given an unordered set $\mathbf{x} = \{x_1, \dots, x_n\} \in \mathcal{F}^n$ such as fingerprint minutiae, if secret, $s \in \mathcal{F}^k$is encoded as coefficients of a $(k-1)$-th order polynomial $f(\cdot)$, then a genuine set $\mathbf{G} = \{(x_1, f(x_1)), \dots, (x_n, f(x_n))\}$ can be generated. The fuzzy vault protects the polynomial $f$ by concealing $\mathbf{G}$ in a vault $\mathbf{V}$. This can be performed by mixing a number of chaff entries $r$, under a chaff set $\mathbf{C} = \{(x_{c(1)}, y_{c(1)}), \dots, (x_{c(r)}, y_{c(r)})\}$ , i.e., $\mathbf{V} = \mathbf{G} \cup \mathbf{C}$ . These chaff entries are to be randomly generated and should not lie on $f$. The chaff set $\mathbf{C}$ is meant to prevent an adversary distinguishing $\mathbf{G}$ from $\mathbf{C}$. For secret retrieval, given a query set $\mathbf{x}' = \{x'_1, \dots, x'_n\} \in \mathcal{F}^n$ , the vault set would be unlocked leading to the revelation of $f$ if $\mathbf{x}'$ is sufficiently close to $\mathbf{x}$ by means of error tolerance. Here, we explain the details of standard polynomial reconstruction via polynomial interpolation over an unlocking set $\mathbf{U} = \{(x_j, f(x_j))\}_{j=1}^{\omega} \subseteq \mathbf{V}$ of size $\omega \leq n$. $\mathbf{U}$ can be generated by identifying the overlapped entries of $\mathbf{x}$ and $\mathbf{x}'$ across the vault $\mathbf{V}$ of size $|\mathbf{V}| = n + r$. In practice, $\mathbf{U}$ may also consist of chaff entries due to the errors in the noisy query set $\mathbf{x}'$ . Therefore, the actual number of overlapped entries of $\mathbf{x}$ and $\mathbf{x}'$ is $t \leq \omega$ . If $k < t$ , then redundancy with $t - k > 0$ in the unlocking set $\mathbf{U}$ must be satisfied to enable error correction. In this sense, $t$ also can be deemed as an error tolerance capacity in the context of fuzzy vault.

Ideally, the mixing of genuine and chaff sets should be uniformly random for ensuring maximum security. However, this assumption does not hold in practice. Over the years, several attempts have been made to rectify this very issue. For instance, Clancy et al. [7] devised a minimum distant criterion away from the genuine set to place the chaff set randomly to eliminate overlapping. This approach was later criticized that the chaff set can be easily recognized and would be prone to a statistical attack, which has a lower complexity than the brute-force attack [8]. Li et al. [9] suggested generating the chaff set in accordance with a minutia descriptor's distribution so that a more "natural" chaff placement can be made. However, this indeed contradicts the uniformity principle of a genuine chaff-set mixture. Moreover, Merkle et al. [10] observed that even the chaff set follows a minutia descriptor's distribution; the chaff set is unlikely to reside in those regions that are frequently occupied by the genuine set. The use of additional information such as the minutia point's orientation to generate the chaff set has been proposed by Nandakumar et al. [11]. However, another study [12] reveals that the minutia points in close proximity tend to have a similar orientation, facilitating the adversary to filter out the chaff set from the genuine set.

Another major concern of practical fuzzy vaults (and all other biometric cryptosystems) is the secret retrieval rate (resilience), which is closely associated with the biometric feature quality. In particular, there exists a failure circumstance such as false-reject despite a genuine feature being input due to large intraclass variations i.e. a query set overlapping with a chaff set. As the failure rate of a single decoding step via polynomial interpolation can be measured as $1 - \binom{\omega}{k} / \binom{|\mathbf{V}|}{k}$, we observe that $\binom{\omega}{k} / \binom{|\mathbf{V}|}{k} > \left(1 - \frac{|\mathbf{V}| - \omega}{|\mathbf{V}| - k}\right)^k \approx e^{-\left(\frac{k(|\mathbf{V}| - \omega)}{|\mathbf{V}| - k}\right)} = \frac{1}{\lambda(k)}$ , and one can expect the failure rate to be negligible after $\lambda(k)$ iterations (proportional to $k$). That is, after repeatedly decoding over randomly selected $k$ unlocking pairs $(x_j, f(x_j)\}$ from the unlocking set $\mathbf{U}$. The failure rate can be minimized by repeating such a decoding process. One typical instance is iterative Lagrange interpolation [13]. However, the number of iterations required by the iterative Lagrange interpolation is impractically large for a huge feature size. One can further reduce the number of iterations required by using more efficient decoding algorithm, i.e., Guruswami Sudan decoder [14], however, it still requires $2^9$ iterations [15] for decoding. In addition, these techniques are restricted by the error tolerance capacity $t$ due to the decoding algorithm that employed. For instance, the correctable error number for Guruswami Sudan decoder [14] is bounded as $n - \sqrt{n(k-1)}$ , which requires $t \geq n - \sqrt{n(k-1)}$. Apart from this, a stronger bound can be derived as follows: with $n$ minutia points and a polynomial of order $k - 1$, the number of errors that can be corrected after $\lambda(k)$ iterations, in principle, is bounded as $n - t \leq k$. This suggests that $\mathbf{x} = \{x_1, \dots, x_n\}$ and $\mathbf{x}' = \{x'_1, \dots, x'_n\}$ must overlap at least 50%, which does not seem realistic for fingerprint minutiae [16].

The brute-force security of fuzzy vault, another key concern, is attributed to the hardness of polynomial reconstruction. This is related to the list-decoding problem of enumerating valid solutions over specific overlapping requirement $(t)$ in between $\mathbf{x}$ and $\mathbf{x}'$ (see [17] for a detailed survey). Recall that the failure rate can be expressed as $1 - \binom{\omega}{k} / \binom{|\mathbf{V}|}{k}$ and $t \leq \omega$. Hence, the failure rate can be used to characterize the brute-force complexity of fuzzy vault as $\binom{|\mathbf{V}|}{k} / \binom{\omega}{k}$ [18]. Notably, the brute-force complexity measures how many iterations are required for an adversary to reconstruct the polynomial successfully. For instance, Nandakumar et al. [11] and Li et al. [9] reported their realizations with brute-force complexity around $2^{32}$ and $2^{35}$, respectively, which are considerably low in practice. Mihailescu [18] pointed out that the practical fuzzy vaults are prone to low brute-force security. This could be due to the limited number of chaff entries that can be added ($r$), the limited number of minutia points ($n$) available, and the polynomial order ($k-1$) constraint.

### *B. Motivations and Contributions*

In this work, we perceive the biometric secret-binding problem as a *symmetric encryption–decryption problem*. In light of this, we devise SKE that comprises Shamir's secret-sharing scheme, RV pairs, and a simple filtering mechanism. In our disposition, an RV pair resembles a symmetric key pair in symmetric cryptosystems and is realized by the respective hashed enrolled and query biometric vectors. The hashing is carried out by IoM hashing with a niche property that preserves the similarity between different inputs [19].

Though both fuzzy vault and SKE use a polynomial to bind a secret, the genuine set of SKE (analog to the *plain text* in symmetric key systems) is encrypted by the RV unlike a chaff set that is required for genuine set concealment in fuzzy vault. Hence, secret retrieval is merely a reverse operation i.e. decryption by means of query RV followed by the polynomial interpolation.

As discussed in the previous section, the resilience is another major concern of biometric cryptosystems due to the noisy nature of biometrics. In our scheme, we show that strong resilience of SKE can be achieved with an overwhelming probability for a genuine biometric input, merely with the IoM hashing and a simple filtering mechanism during decryption (Section IV).

Though SKE is a biometric cryptosystem, it inherits the legacy of symmetric key cryptosystems. Hence, its threat model includes that from classical biometric cryptosystems such as *brute-force attack* and *false-accept attack*. Besides, we consider another notion called *tag indistinguishability* adopted from the concept of key privacy or anonymity (unable to differentiate which key is being used for encryption) that is often used in public key cryptosystems [20] and identity-based cryptosystems [21]. In our context, as a biometric input is unique and is lifetime associated to an individual, the notion of tag indistinguishability offers strong privacy protection to any user to remain anonymous while encryption is exercised by SKE. We will analyze them formally and comprehensively in Section V.

To sum up, the contributions of this paper are four-fold:

1. We put forward a novel simplistic biometric secret-binding scheme for vectorial biometrics, namely SKE that is based on the notion of symmetric key cryptosystems.
2. We demonstrate the innovative use of IoM hashing that allows one to generate *abundant* IoM hashed entries as genuine entries in SKE without being restricted by the original biometric vector size. Hence, the failure probability of secret retrieval via Shamir's secret-sharing mechanism due to corrupted genuine entries can be reduced radically. Besides, this unique trait allows us to choose a *higher order polynomial*, which is beneficial to strengthen the brute-force and false-accept securities.
3. We formalize and analyze the threat model of SKE that consists of three major security threats, namely brute-force attack, false-accept attack, and tag indistinguishability and show that SKE resists these attacks.
4. Though SKE was motivated as an error-tolerant symmetric encryption construct, it serves as a biometric template protection (BTP) scheme as well [22] [23]. Therefore, we show that SKE satisfies the four design criteria of BTP, namely noninvertibility, unlinkability, revocability, and performance (Section VII).

Although SKE is generic for any vectorial biometrics, we showcase the realization of SKE with fingerprint vectors [24]. We show the empirical results comprehensively with five subsets of fingerprint benchmarks under FVC 2002, FVC 2004, and FVC 2006.

The MATLAB code of SKE is available at (goo.gl/8EoLsp), allowing researchers to reproduce or verify the results of this study.

## II. PRELIMINARIES

In this section, we give a brief introduction about Rivest's keyring model as well as IoM hashing.

### *A. Rivest's Keyring Model*

The keyring model is an error-tolerant symmetric cryptosystem proposed by Rivest [6]. The keyring refers to a "bag of keywords" that is distributed in a pair to both sender and receiver. This model relies on the keyword-matching game, which is favorable for two similar keyrings. Each keyring can be represented as a binary vector where each single bit "1" corresponds to a particular keyword in the universe set $\mathcal{U}$. For Alice and Bob who have similar keyrings $\mathbf{x} \in \mathcal{U}$ and $\mathbf{x}' \in \mathcal{U}$ respectively, random vectors with length $m$ such as $\boldsymbol{\phi}_{\mathbf{x}} = \{\varphi_{\mathbf{x}(1)}, \dots, \varphi_{\mathbf{x}(m)}\} \in \mathcal{U}^m$ and $\boldsymbol{\phi}_{\mathbf{x}'} = \{\varphi_{\mathbf{x}'(1)}, \dots, \varphi_{\mathbf{x}'(m)}\} \in \mathcal{U}^m$ can be generated through their respective *resilient set vectorizer* (RSV), $\Omega(\mathbf{x}, m, \mathbf{N}) \rightarrow \boldsymbol{\phi}_{\mathbf{x}}$ and $\Omega(\mathbf{x}', m, \mathbf{N}) \rightarrow \boldsymbol{\phi}_{\mathbf{x}'}$ with a shared nonce $\mathbf{N}$ (e.g., public random numbers). The random vector pair $\{\boldsymbol{\phi}_{\mathbf{x}}, \boldsymbol{\phi}_{\mathbf{x}'}\}$ can then be employed to encrypt or decrypt a secret along with an ECC.

Formally, a RSV is a set vectorizer with length $m$, along with the property that for any two keyrings $\mathbf{x}$ and $\mathbf{x}'$ with $P = \mathrm{J}(\mathbf{x}, \mathbf{x}') = |\mathbf{x} \cap \mathbf{x}'|/|\mathbf{x} \cup \mathbf{x}'|$ where $\mathrm{J}(.,.)$ is the Jaccard similarity coefficient. It follows $t \sim \mathrm{Bin}(m, P)$ where $t$ is the number of positions wherein $\boldsymbol{\phi}_{\mathbf{x}}$ and $\boldsymbol{\phi}_{\mathbf{x}'}$ agree. More explicitly, if a fraction $P$ of $\mathbf{x} \cup \mathbf{x}'$ is shared, then the fraction of the positions where $\boldsymbol{\phi}_{\mathbf{x}}$ and $\boldsymbol{\phi}_{\mathbf{x}'}$ agree follows a binomial distribution parameterized by $m$ and $P$. In contrast, the number of disagree positions can be characterized by the hamming weight of the offset $||\boldsymbol{\delta}|| = ||\boldsymbol{\phi}_{\mathbf{x}} \oplus \boldsymbol{\phi}_{\mathbf{x}'}||$. Its expected value is $\mathbb{E}(||\boldsymbol{\delta}||) = m(1 - \mathrm{J}(\mathbf{x}, \mathbf{x}'))$. In Rivest's proposal, he stated a way of playing the keyword-matching game by means of min-hashing [25] as the resilient set vectorizer.

Rivest's keyring model resembles fuzzy commitment in such a sense that Alice conceals a codeword $\mathbf{c} = \{c_1, \dots, c_m\} \in \mathcal{U}^m$ with $\boldsymbol{\phi}_{\mathbf{x}}$ via secure sketch $\mathbf{ss} = \mathbf{c} \oplus \boldsymbol{\phi}_{\mathbf{x}} = \{c_1 \oplus \varphi_{\mathbf{x}(1)}, \dots, c_m \oplus \varphi_{\mathbf{x}(m)}\}$. Then, Bob would be able to identify the nearest codeword $\mathbf{c}'$ in a reverse manner such as $\mathbf{c}' = \mathbf{ss} \oplus \boldsymbol{\phi}_{\mathbf{x}'} = \mathbf{c} \oplus \boldsymbol{\phi}_{\mathbf{x}} \oplus \boldsymbol{\phi}_{\mathbf{x}'} = \boldsymbol{\delta} \oplus \mathbf{c}$. With a suitable ECC over tolerance capacity $t$, Bob is expected to retrieve the secret with a high probability if $t > \mathbb{E}(||\boldsymbol{\delta}||) = m\big(1 - \mathrm{J}(\mathbf{x}, \mathbf{x}')\big)$.

### *B. Index-of-Max Hashing*

IoM hashing [19] is a special instance of locality-sensitive hashing (LSH) [26] that consumes feature vector $\mathbf{x} \in \mathbb{R}^d$ over continuous domain $\mathbb{R}^d$. Let $\mathbf{Y} \in \mathbb{R}^{\mathcal{F} \times d}$ be a random projection matrix composed of $\mathcal{F}$ Gaussian row vectors where each follows a standard normal $\mathcal{N}(0,1)$ distribution. Further, let $\mathbf{N} = \{\mathbf{Y}_1, \ldots, \mathbf{Y}_m\}$ be an independent random projection matrix set.

Given $\mathbf{x}$, $m$, and $\mathbf{N}$, the IoM hashing is defined as a mapping function $\Omega^{\mathrm{IoM}}(\mathbf{x}, m, \mathbf{N}): \mathbb{R}^{\mathcal{F} \times d} \times \mathbb{R}^d \times m \to \mathcal{F}^m$ . The operation of $\Omega^{\mathrm{IoM}}(\mathbf{x}, m, \mathbf{N})$ is given as follows:

1. Record the *indices* of the maximum value computed from $\varphi_{\mathbf{x}(i)} = \arg\max_i \langle \mathbf{Y}_i, \mathbf{x} \rangle \in \mathcal{F}$ , where $\langle , \rangle$ is the inner product.
2. Repeat step 1 for $i = 1, \ldots, m$ and yield $\boldsymbol{\phi}_{\mathbf{x}} = [\varphi_{\mathbf{x}(1)}, \ldots, \varphi_{\mathbf{x}(m)}]$.

In other words, the IoM hashing [19] embeds $\mathbf{x}$ onto a $\mathcal{F}$-dimensional Gaussian random subspace to output a random integer over a field of size $\mathcal{F}$ corresponding to the index of the maximum value. This process is repeated with $m$ sets of independent Gaussian random matrices and random integer vectors to yield a collection of $m$ independent IoM entries, $\varphi_{\mathbf{x}(i)} \in \{0,1, \ldots \mathcal{F} - 1\}$, for $i = 1, \ldots, m$.

Both min hashing (suggested in Rivest's keyring model) and IoM hashing essentially follow the LSH spirit [26], and both strive to ensure that the two vectors $\mathbf{x}$ and $\mathbf{x}'$ with high similarity render a higher probability of collision (number of agreed entries) between their output vectors $\boldsymbol{\phi}_{\mathbf{x}}$, and $\boldsymbol{\phi}_{\mathbf{x}'}$ respectively. On the contrary, if $\mathbf{x}$ and $\mathbf{x}'$ are far apart, it will result in a low probability of hash collision between $\boldsymbol{\phi}_{\mathbf{x}}$, and $\boldsymbol{\phi}_{\mathbf{x}'}$.

Specifically, let $\boldsymbol{\phi}_{\mathbf{x}} = [\varphi_{\mathbf{x}(1)}, \ldots, \varphi_{\mathbf{x}(m)}]$ and $\boldsymbol{\phi}_{\mathbf{x}'} = [\varphi_{\mathbf{x}'(1)}, \ldots, \varphi_{\mathbf{x}'(m)}]$ be the enrolled and query IoM hashed vectors, respectively. We can calculate the similarity of $\boldsymbol{\phi}_{\mathbf{x}}$ and $\boldsymbol{\phi}_{\mathbf{x}'}$ by measuring their number of collisions $t = \sum_{i=1}^{m} X_i$, where $X_i \in \{0,1\}$ is an independent and identically distributed (i.i.d) variable such that $X_i = 1$ if $\varphi_{\mathbf{x}(i)} = \varphi_{\mathbf{x}'(i)}$ and 0 otherwise. Formally, one has $\mathbb{P}(X_i = 1) = \mathbb{P}(\varphi_{\mathbf{x}(i)} = \varphi_{\mathbf{x}'(i)}) = S(\mathbf{x}, \mathbf{x}')$, where $S(\mathbf{x}, \mathbf{x}') = \frac{1}{\mathcal{F}} + \sum_{j=1}^{\infty} a_j(\mathcal{F})(\cos\theta)^i$ is defined as the similarity function of IoM hashing with $\cos\theta = \frac{\mathbf{x} \cdot \mathbf{x}'}{\|\mathbf{x}\|\|\mathbf{x}'\|}$ and $a_j(\mathcal{F})$ is the coefficient that satisfies $\frac{1}{\mathcal{F}} + \sum_{j=1}^{\infty} a_j(\mathcal{F}) = 1$ [19].

In this paper, the IoM hashing $\Omega^{\mathrm{IoM}}(\mathbf{x}, m, \mathbf{N})$ is portrayed as an instance of RSV where $\mathbf{N}$ corresponds to the public nonce in Rivest's Keyring model and $t \sim \mathrm{Bin}\big(m,\ S(\mathbf{x}, \mathbf{x}')\big)$ with $\mathbb{E}(t) = mS(\mathbf{x}, \mathbf{x}')$.

## III. Symmetric Keyring Encryption

SKE inherits the RSV notion from the keyring model as a means of symmetric key pair generation. Specifically, IoM hashing is naturally adopted by SKE as RSV as presented in Section IIB. In what follows, the public random Gaussian matrices set in IoM hashing and biometrics in the vector form respectively correspond to nonce and keyring in the keyring model. We refer different nonce $\mathbf{N}$ and $\widehat{\mathbf{N}}$ to different random Gaussian matrix sets. These nonce $\mathbf{N}$ and $\widehat{\mathbf{N}}$ are used to generate different random vectors $\boldsymbol{\phi}_{\mathbf{x}} \in \mathcal{F}^m$ and $\widehat{\boldsymbol{\phi}}_{\mathbf{x}} \in \mathcal{F}^m$ respectively, called RVs. The main idea of SKE is as follows.

### *A. Main Idea of SKE*

Given that SKE is primarily used for secret binding and retrieval, we adopt Shamir's threshold secret-sharing scheme for this purpose. Here, a finite field polynomial $f(.) \in \mathcal{F}$ of order $k - 1$ is used for secret embedding.

During enrollment, a user with an input biometric vector $\mathbf{x}$ first generates an RV $\boldsymbol{\phi}_{\mathbf{x}}$ through RSV with nonce $\mathbf{N}$, together with its polynomial projected correspondences, $f(\boldsymbol{\phi}_{\mathbf{x}})$. A genuine set $\mathbf{G} = \Big\{\Big(\varphi_{\mathbf{x}(1)}, f\big(\varphi_{\mathbf{x}(1)}\big)\Big), \ldots, \Big(\varphi_{\mathbf{x}(m)}, f\big(\varphi_{\mathbf{x}(m)}\big)\Big)\Big\}$ is formed based on $\boldsymbol{\phi}_{\mathbf{x}}$ and $f(\boldsymbol{\phi}_{\mathbf{x}})$. At this point, SKE resembles fuzzy vault where $\mathbf{G}$ is used to bind a secret over the coefficients of $f$. However, instead of concealing $\mathbf{G}$ with a chaff set, we merely hide $f(\boldsymbol{\phi}_{\mathbf{x}})$. This is done via a second RV $\widehat{\boldsymbol{\phi}}_{\mathbf{x}} = \{\widehat{\varphi}_{\mathbf{x}(1)}, \ldots, \widehat{\varphi}_{\mathbf{x}(m)}\}$ with nonce $\widehat{\mathbf{N}}$. The second RV acts as a random "keyset," which is independent of $\boldsymbol{\phi}_{\mathbf{x}}$ to encrypt $f\big(\varphi_{\mathbf{x}(i)}\big)$, $i$ = 1,…,$m$ and yields a public secure sketch $\mathbf{ss} = \{ss_1, \ldots, ss_m\}$, where $ss_i = \widehat{\varphi}_{\mathbf{x}(i)} \oplus f\big(\varphi_{\mathbf{x}(i)}\big)$. As such, the encryption simply follows fuzzy commitment with an XOR operation yet accompanies with an authentication tag $\mathbf{TAG} = \{\mathrm{tag}_1, \ldots, \mathrm{tag}_m\}$, where $\mathrm{tag}_i = H\big(\varphi_{\mathbf{x}(i)} || ss_i || \widehat{\varphi}_{\mathbf{x}(i)}\big)$ and $H$ is a cryptographic secure one-way hash function. Note that $||$ refers to concatenation.

During secret retrieval, a query RV pair $\big(\boldsymbol{\phi}_{\mathbf{x}'}, \widehat{\boldsymbol{\phi}}_{\mathbf{x}'}\big)$ can be generated with nonce $(\mathbf{N}, \widehat{\mathbf{N}})$ from the query biometric vector $\mathbf{x}'$. By doing so, one can verify the integrity of the secure sketch along with the authentication $\mathrm{tag}_i$ empowered filtering mechanism. To be specific, as $ss_i = f\big(\varphi_{\mathbf{x}(i)}\big) \oplus \widehat{\varphi}_{\mathbf{x}(i)}$ and if all the three conditions are satisfied i.e. $\varphi_{\mathbf{x}(i)} = \varphi_{\mathbf{x}'(i)}$, $\widehat{\varphi}_{\mathbf{x}(i)} = \widehat{\varphi}_{\mathbf{x}'(i)}$, and $ss_i$ remain unaltered, then one can generate $\mathrm{tag}'_i = H\big(\varphi_{\mathbf{x}'(i)} || ss_i || \widehat{\varphi}_{\mathbf{x}'(i)}\big)$ that appears identical to $\mathrm{tag}_i$. Therefore, the decryption succeeds and yields $f'\big(\varphi_{\mathbf{x}(i)}\big) = ss_i \oplus \widehat{\varphi}_{\mathbf{x}'(i)} = f\big(\varphi_{\mathbf{x}(i)}\big)$. On the other hand, if not all the three conditions hold, then $f'\big(\varphi_{\mathbf{x}(i)}\big) \neq f\big(\varphi_{\mathbf{x}(i)}\big)$ and decryption fails. From this perspective, when $\mathrm{tag}'_i = \mathrm{tag}_i$ , a genuine pair $\Big(\varphi_{\mathbf{x}(i)}, f\big(\varphi_{\mathbf{x}(i)}\big)\Big) \in \mathbf{G}$ can be revealed subject to $H()$ must satisfy the collision resistant property. For a sufficient number of revealed genuine pairs, one can construct an unlocking set $\mathbf{U} = \Big\{\Big(\varphi_{\mathbf{x}(j)}, f\big(\varphi_{\mathbf{x}(j)}\big)\Big)\Big\}_{j=1}^{t} \subseteq \mathbf{G}$. If $t \geq k$, the secret can be retrieved via polynomial interpolation using $\mathbf{U}$. The high-level overview of SKE is illustrated in Figure 1. The detailed steps of enrollment and secret retrieval will be given in the following subsections.

### *B. Enrollment*

Given input $\mathbf{x} \in \mathbb{R}^d$, two different nonce $\mathbf{N}$, $\widehat{\mathbf{N}}$, parameter $m$, a finite field polynomial with degree $k - 1$ $f(.)$ , and a one-way hash function $H : \{0,1\}^* \to \{0,1\}^\ell$ , the enrollment procedure of SKE is as follows:

1. Generate $\boldsymbol{\phi}_{\mathbf{x}} \leftarrow \Omega^{\mathrm{IoM}}(\mathbf{x}, m, \mathbf{N})$ and $\widehat{\boldsymbol{\phi}}_{\mathbf{x}} \leftarrow \Omega^{\mathrm{IoM}}(\mathbf{x}, m, \widehat{\mathbf{N}})$, where $\boldsymbol{\phi}_{\mathbf{x}} = \{\varphi_{\mathbf{x}(1)}, \dots, \varphi_{\mathbf{x}(m)}\}$ and $\widehat{\boldsymbol{\phi}}_{\mathbf{x}} = \{\widehat{\varphi}_{\mathbf{x}(1)}, \dots, \widehat{\varphi}_{\mathbf{x}(m)}\}$.
2. Perform polynomial projection to generate $f(\boldsymbol{\phi}_{\mathbf{x}}) = \{f(\varphi_{\mathbf{x}(1)}), \dots, f(\varphi_{\mathbf{x}(m)})\}$.
3. Encrypt $f(\boldsymbol{\phi}_{\mathbf{x}})$ to yield secure sketch $\mathbf{ss} = \{\widehat{\varphi}_{\mathbf{x}(1)} \oplus f(\varphi_{\mathbf{x}(1)}), \dots, \widehat{\varphi}_{\mathbf{x}(m)} \oplus f(\varphi_{\mathbf{x}(m)})\}$.
4. Generate authentication tag $\mathbf{TAG} = \{\mathrm{tag}_1, \dots, \mathrm{tag}_m\}$ where $\mathrm{tag}_i = H(\varphi_{\mathbf{x}(i)} || ss_i || \widehat{\varphi}_{\mathbf{x}(i)})$ is the one-way hashed output.
5. Store $\{\mathbf{N}, \widehat{\mathbf{N}}, \mathbf{TAG}, \boldsymbol{ss}\}$ as the public helper data.

*C. Secret Retrieval*

Given $\mathbf{x}' \in \mathbb{R}^d$ the query biometric vector and the public helper data $\{\mathbf{N}, \widehat{\mathbf{N}}, \mathbf{TAG}, \mathbf{ss}\}$ with input parameter $m$ and the same one-way hashing $H$, the detailed key retrieval steps are as follows:

1. Generate $\boldsymbol{\phi}_{\mathbf{x}'} \leftarrow \Omega^{\mathrm{IoM}}(\mathbf{x}', m, \mathbf{N})$ and $\widehat{\boldsymbol{\phi}}_{\mathbf{x}'} \leftarrow \Omega^{\mathrm{IoM}}(\mathbf{x}', m, \widehat{\mathbf{N}})$ , where $\boldsymbol{\phi}_{\mathbf{x}'} = \{\varphi_{\mathbf{x}'(1)}, \dots, \varphi_{\mathbf{x}'(m)}\}$ and $\widehat{\boldsymbol{\phi}}_{\mathbf{x}'} = \{\widehat{\varphi}_{\mathbf{x}'(1)}, \dots, \widehat{\varphi}_{\mathbf{x}'(m)}\}$.
2. Compute $\mathbf{TAG}' = \{\mathrm{tag}'_1, \dots, \mathrm{tag}'_m\}$ , where $\mathrm{tag}'_i = \mathrm{H}(\varphi_{\mathbf{x}'(i)} || ss_i || \widehat{\varphi}_{\mathbf{x}'(i)})$.
3. Run Algorithm 1 with input $\{\mathbf{TAG}, \mathbf{TAG}', \boldsymbol{\phi}_{\mathbf{x}}, \boldsymbol{\phi}_{\mathbf{x}'}, \mathbf{ss}\}$ and yield unlocking set $\mathbf{U} = \left\{\left(\varphi_{\mathbf{x}(j)}, f(\varphi_{\mathbf{x}(j)})\right)\right\}_{j=1}^{t} \subseteq \mathbf{G}$.
4. Perform polynomial reconstruction with $\mathbf{U}$.

In step 4, the secret can be retrieved via Lagrange interpolation if $t \geq k$. Given that we have $\left(\varphi_{\mathbf{x}(i)}, f(\varphi_{\mathbf{x}(i)})\right) \in \mathcal{F} \times \mathcal{F}$, this means there will be at most $\mathcal{F} \times \mathcal{F}$ numbers of unique genuine pairs in $\mathbf{U}$. Besides, the choice of $\mathcal{F}$ and $m$ is indeed interrelated where $\mathcal{F} < m$ is often opted to realize improved accuracy performance [19]. This implies that the key retrieval rate could be degraded if one chooses to increase $\mathcal{F}$ close or equal to $m$. As a solution, a large value of $m$ is needed to compensate for the increment of $\mathcal{F}$. Consequently, this may lead to more nonunique genuine pairs in $\mathbf{U}$ and increase the computation complexity of polynomial interpolation. Therefore, filtering out these nonunique genuine pairs as detailed in Algorithm 1 is crucial. Note that $\mathrm{Unique}(\mathbf{U})$ is a function that warrants only *unique genuine pairs* present in $\mathbf{U}$.

The use of **TAG** in SKE is indispensable as it serves three purposes: (1) it performs a conditional check to ensure only relevant $ss_i$ i.e., when $\mathrm{tag}_i = \mathrm{tag}'_i$, can be decrypted. This can be seen as an error-checking mechanism for $\varphi_{\mathbf{x}(i)}$ and $\varphi_{\mathbf{x}'(i)}$ as well; (2) data integrity of $\boldsymbol{ss}$ is ensured by **TAG**, as if any alteration occurs in $ss_i$, the decryption would fail. (3) The use of TAG facilitates polynomial interpolation by ensuring that only genuine pairs are within $\mathbf{U}$, and hence secret can be retrieved simply with a single step of polynomial reconstruction rather than iterative decoding.

It is useful to illustrate SKE using an example with numerical explication. Let **x** and **x'** be the enrolled and query biometric vectors of Alice, respectively. **x** and **x'** are not identical due to noise but close.

**Example 1**

Public Algorithm: $H$, Algorithm 1

Public Parameters**:** $m = 5$, $k = 3$, $\mathcal{F} = 5$

**Enrollment:**

1. Given a secret $s = 2$, Alice encodes the secret to a finite field polynomial order $k - 1$, i.e. $f = x^2 + 2x + s \bmod (\mathcal{F})$. Suppose she generates $\boldsymbol{\phi}_{\mathrm{x}} = \{3, 1, 2, 3, 2\} \in \mathcal{F}^m$ and $\widehat{\boldsymbol{\phi}}_{\mathbf{x}} = \{2, 3, 0, 1, 1\} \in \mathcal{F}^m$ with her biometric vector $\mathbf{x}$ via $\mathbf{N}$, and $\widehat{\mathbf{N}}$, respectively.
2. Alice performs polynomial projection $(\boldsymbol{\phi}_{\mathbf{x}}) = \{f(3), f(1), f(2), f(3), f(2)\} = \{2, 2, 1, 2, 1\}$.
3. Alice encrypts $f(\boldsymbol{\phi}_{\mathbf{x}})$ via $\widehat{\boldsymbol{\phi}}_{\mathbf{x}}$ to generate the secure sketch $\mathbf{ss}$, that is $\mathbf{ss} = \{10, 10, 01, 10, 01\} \oplus \{10, 11, 00, 01, 01\}$ = $\{00, 01, 10, 11, 00\}$ = $\{0, 1, 2, 3, 0\}$. She generates $\mathbf{TAG}$ = $\{H(3||0||2)$ , $H(1||1||3)$ , $H(2||2||0)$ , $H(3||3||1)$ , $H(2||0||1)\}$.
4. Alice stores $\{\mathbf{N}, \widehat{\mathbf{N}}, \mathbf{TAG}, ss\}$ .

| Algorithm 1: Filtering mechanism for Error Checking |
|---|
| Input: $\mathbf{TAG}, \mathbf{TAG}', \boldsymbol{\phi}'_{\mathrm{x}}, \boldsymbol{\phi}'_{\mathrm{RSV}}, \mathbf{ss}$ |
| $\mathbf{U} \leftarrow \emptyset$<br>**For** $i = 1, \dots, m$<br>    **If** $(\mathrm{TAG}_i = \mathrm{TAG}'_i)$<br>        $f(\varphi_{\mathbf{x}(i)})' = ss_i \oplus \widehat{\varphi}_{\mathbf{x}'(i)}$<br>$\mathbf{U} \leftarrow \mathbf{U} \cup \left(\varphi_{\mathbf{x}'(i)}, f(\varphi_{\mathbf{x}(i)})'\right)$<br>    **End if**<br>**End for**<br>$\mathbf{U} \leftarrow \mathrm{Unique}(\mathbf{U})$ |
| Output: $\mathbf{U}$ |

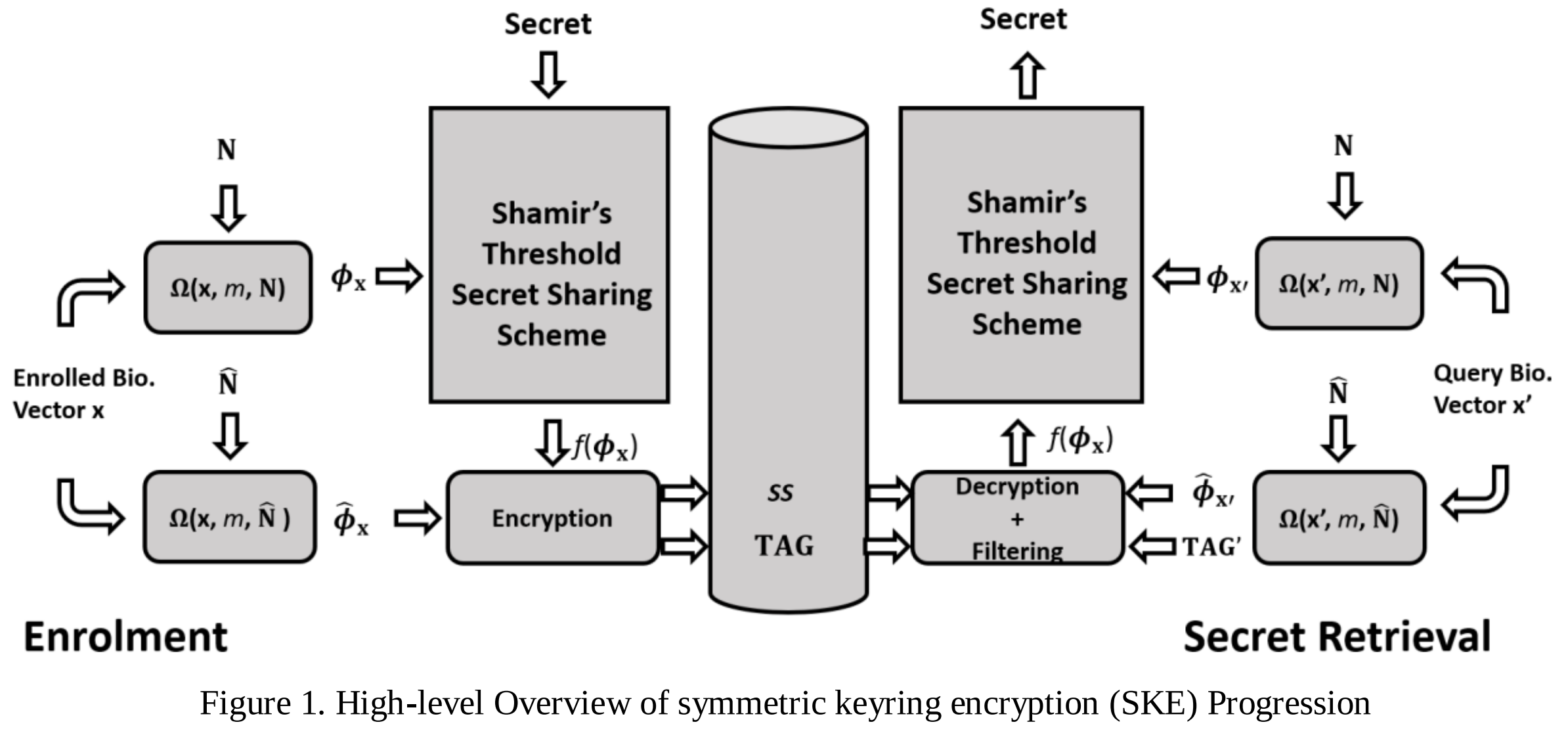

Figure 1. High-level Overview of symmetric keyring encryption (SKE) Progression

**Secret Retrieval:**

1. Alice generates $\boldsymbol{\phi}_{\mathbf{x}'} = \{3, 1, 1, 2, 2\} \in \mathcal{F}^m$ and $\widehat{\boldsymbol{\phi}}_{\mathbf{x}'} = \{2, 3, 1, 0, 1\} \in \mathcal{F}^m$ with $\mathbf{x}'$ via $\mathbf{N}$, and $\widehat{\mathbf{N}}$, respectively.
2. Using the identical one-way hash function, Alice generates $\mathbf{TAG}'$ = $\{H(3||0||2)$,$H(1||1||3)$, $H(1||2||1)$, $H(2||3||0)$, $H(2||0||1)\}$.
3. She runs Algorithm 1 to yield $\mathbf{U}$. Given that three entries of $\mathbf{TAG}$ and $\mathbf{TAG}'$ are matched, i.e. $\{H(3||0||2), H(1||1||3), H(2||0||1)\}$ , this implies that Alice can regenerate $f$ with Lagrange interpolation with the unique genuine pair, i.e. $\mathbf{U} = \{(3, P(3)), (1, P(1)), (2, P(2))\}$ because $|\mathbf{U}| = t \geq k$ . The secret can be reconstructed, and the correctness can be verified through $f(0) = f(0)' = s = 2$.

Note that the choice of $\mathbf{N}$ and $\widehat{\mathbf{N}}$ are random and independent of biometric input though $\mathbf{ss}$ and $\mathbf{TAG}$ are RV or $\mathbf{x}$ and $\mathbf{x}'$ dependent. Therefore, without knowing $\mathbf{x}$ or $\mathbf{x}'$, the knowledge of $\mathbf{N}$ and $\widehat{\mathbf{N}}$ alone does not offer additional advantage in generating the RV, which can reveal the clue for $\mathbf{ss}$ decryption. Figure 2 details the enrollment and secret retrieval operation of SKE.

## IV. Resilience Analysis

In this section, we elaborate the resilience property of SKE. Suppose a pair of RV entries (enrolled and query), $\varphi_{\mathbf{x}(i)}$ and $\varphi_{\mathbf{x}'(i)}$ are generated by $\mathbf{N}$. The same goes to the other pair, $\hat{\varphi}_{\mathbf{x}(i)}$ and $\hat{\varphi}_{\mathbf{x}'(i)}$ generated by $\widehat{\mathbf{N}}$.

Based on the i.i.d attribute of IoM hashed entries, the collision probability of the enrolled and query entries is $\mathbb{P}(\varphi_{\mathbf{x}(i)} = \varphi_{\mathbf{x}'(i)}) = P = S(\mathbf{x}, \mathbf{x}')$ , where $S(\mathbf{x}, \mathbf{x}')$ is the compliance similarity function of IoM hashing given in Section II(B). The same goes to $\mathbb{P}(\hat{\varphi}_{\mathbf{x}(i)} = \hat{\varphi}_{\mathbf{x}'(i)}) = P$. Hence, the joint collision probability of two probabilities is merely their product, i.e. $P^2 = (1 - \text{dis}(\mathbf{x}, \mathbf{x}'))^2$ due to independent $\mathbf{N}$ and $\widehat{\mathbf{N}}$ . In what follows, the number of collision $t$ follows $\text{Bin}(m, P^2)$, where $P^2 = (S(\mathbf{x}, \mathbf{x}'))^2$, expected value $\mathbb{E}(t) = mP^2$, and variance $\text{Var}(t) = mP^2(1 - P^2)$.

For further resilience evaluation, we let $\delta \in (0,1)$ be a fraction of $m$ and $k = \lceil \delta m \rceil$. This allows us to define a security threshold in terms of the fraction of $\delta = \frac{k}{m}$ rather than solely on $k$. Intuitively, for high resilience, given $\mathbf{x} \in \mathbb{R}^d$ to bind a secret $s \in \mathcal{F}^k$ with parameter $(m, \delta)$, one expects that any similar $\mathbf{x}' \in \mathbb{R}^d$ will retrieve the secret with an overwhelming probability (probability close to 1) if $\mathbb{E}(t) = mP^2 > k$ or $\mathbb{E}(t)/m = P^2 > \delta$.

Here we give an example to illustrate the resilience of SKE.

**Example 2**: Consider a polynomial of order $k - 1$ with $k = \lceil \delta m \rceil$. Suppose $m = 256$ and hence exact secret retrieval is only possible if $t > k = \lceil \delta m \rceil = 192$ for $\delta = \frac{3}{4}$. In other words, more than 75% of $\mathbf{TAG}(\mathbf{TAG}')$ entries $\text{tag}_i(\text{tag}'_i)$ need to be successfully verified (e.g.,$\text{tag}_i = \text{tag}'_i$). It is reasonable to assume that for two genuine biometric vector pairs, $S(\mathbf{x}, \mathbf{x}') = 0.9$ (i.e. 90% similar); hence, the probability of success for $\text{tag}_i = \text{tag}'_i$ is $P^2 = S(\mathbf{x}, \mathbf{x}')^2 = 0.81$. As $t \sim \text{Bin}(256,\ 0.81)$ with $\mathbb{E}(t) = 207 > k$ , the secret can be retrieved with $\mathbb{P}(t > 192) = 0.989$, an overwhelming probability for exact secret retrieval.

The above example suggests that to achieve high resilience, one can decrease $\delta$ so that $\mathbb{E}(t)/m = P^2 > \delta$ can be easily satisfied.

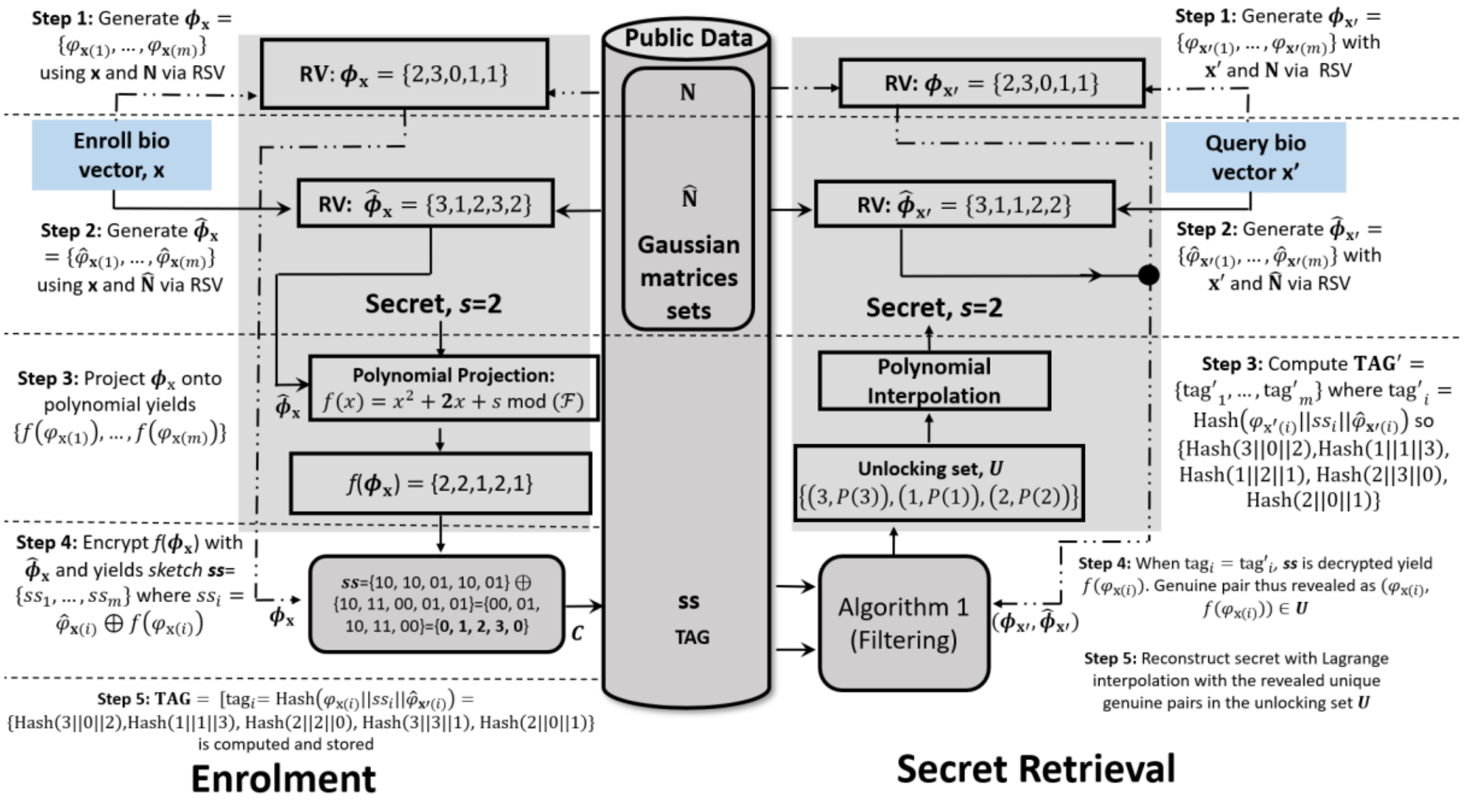

Figure 2: Illustration of secret binding and retrieval in SKE.

## V. Security

In this section, we present the threat model of SKE. We divide the threats of SKE into two categories. First, as SKE is essentially a biometric cryptosystem, conventional security attacks such as brute-force attack and false-accept attack will be discussed. Second, because SKE is a special kind of error-tolerant symmetric key cryptosystem that incorporates the tag, we shall analyze SKE security in terms of tag indistinguishability.

### A. Security Model

Before we elaborate each potential security threat, we formalize the security model of SKE. Typically, it is straightforward to reduce our security to the computational hardness in finding $k$ out of $m$ collisions on $\mathbf{TAG}$ subject to $\text{tag}_i = \text{tag}'_i$, $i = 1, \ldots, m$ by using a ROM over a uniformly random query vector. This security can be claimed by asserting higher values for $\mathcal{F}$ and $m$, because doing so can increase the difficulty for one to launch a brute-force attack using the same one-way hash function $H$ under the ROM. However, this statement is not sufficient for our security goal, which is to show that given a query biometric vector $\mathbf{x}'$, it is infeasible to retrieve the bound secret unless $\mathbf{x}'$ is close to $\mathbf{x}$ without assuming that $\mathbf{x}'$ is uniformly random. To resolve this issue, the security goal is preferably characterized by the similarity of $\mathbf{x}$ and $\mathbf{x}'$, which is more naturally distributed over some random distribution, not necessary to be uniform. In light of this, we exploit the input structure of the biometric vectors by means of their similarity using RSV. Formally, we adopt a *random error model* (See Section 8 in [5], and [27] for similar realization), whereby different entries of the RV pair come with a probability of colliding i.e. $\text{tag}_i = \text{tag}'_i$ characterized as $P^2$, and noncolliding (errors) given as $1 - P^2$, where $P^2 = S(\mathbf{x}, \mathbf{x}')^2$ (Section II.B). This permits us to show security that is governed by the nonuniform biometric vectors and has high resilience can be achieved in accordance to their similarity. More importantly, our security property holds under the standard model without relying on the ROM, but only requires that $H$ is collision-resistant.

However, it is unrealistic to assume that the error process of the biometric vector can be modeled accurately, and same goes to collision probability $P$ estimation. Therefore, meaningful security needs to have a precise knowledge of the input distribution, and this demands considerable human effort and collection of a massive number of biometric samples, which is expensive process. A notable example is the collection of IrisCode by Daugman [28], and claims the IrisCode security by argument on the *degree of freedom* over 11.5 million of IrisCode samples. We attempt to relax the security requirement due to limited samples and therefore use the *maximum collision probability* denoted as $P_{M_I}^{\max}$. Formally, $P_{M_I}^{\max} = \max\{P_1, \ldots, P_{M_I}\}$ is, in principle, the maximum imposter matching score of the RV pairs over $M_I$ scores of different query biometric vectors $\mathbf{x}'$. This enables us to demonstrate security over a smaller sample size that considered the inherited nonuniform nature of the biometric vector.

### B. Brute-force Attack

We first argue on the brute-force attack. A brute-force attack for SKE refers to a circumstance when an adversary attempts to guess the secret directly.

Let the secret in the finite field be $s \in \mathcal{F}^k$, where $k$ is the polynomial order. The complexity of this attack relies on the secret space size i.e. $\mathcal{F}^k$. If a secret is encoded in 8 bits, $|\mathcal{F}| = 2^8$, and $k = 20$ is chosen, then the Brute-force (BF) complexity of SKE is $2^{160}$ to guess all the polynomial coefficients.

### C. False-accept Attack

False-accept attack (also known as dictionary attack) refers to a scenario where an adversary manages to compromise a large dataset. He/she can repeatedly attempt to verify the $\mathbf{TAG}$ with the compromised biometric instances. In this circumstance, the adversary is expected to gain access to the system with the probability equal to the nonzero false-accept rate (FAR) of the system. As this attack exploits the FAR of the biometric system, it is possible to use an artificial template generator to launch this attack without the need for compromising any dataset. Besides, it is also feasible for any adversary with a higher computation power to have better modeling on the input biometric distribution, allowing the adversary to launch the false-accept attack by randomly sampling a fingerprint biometric vector according to the specific distribution. From this point of view, the false-accept attack complexity can be defined in terms of *false-acceptance probability* that can be calculated in the same way as shown in Example 2 with $P^2 = \left(P_{M_I}^{\max}\right)^2$. Therefore, we have the *maximum false-accept attack complexity* over $M_I$ secret retrieval attempts described as

$$\mathbf{fa}\left(m, \delta, P_{M_I}^{\max}\right) = -\log\left(\mathbb{P}\left(t_{\left(P_{M_I}^{\max}\right)} \geq \lceil \delta m \rceil\right)\right), \quad (1)$$

where $t_{(P_M^{\max})} \sim \text{Bin}(m, (P_{M_I}^{\max})^2)$.

Clearly, unlike the brute-force attack, the false-accept attack is independent of $\mathcal{F}$. Therefore, increasing $\mathcal{F}$ would not contribute higher false-accept security, and thus we deem the false-accept attack is a much stronger attack that could serve as the lower bound to brute-force attack through a considerably large $\mathcal{F}$ could be supplied in practice.

Apart from this, we also can reason the complexity of the false-accept attack as follows. Given a $\text{tag}_i = H\left(\varphi_{\mathrm{x}(i)} || ss_i || \hat{\varphi}_{\mathrm{x}(i)}\right)$, where $\varphi_{\mathrm{x}(i)} \in \mathcal{F}$ and $\hat{\varphi}_{\mathrm{x}(i)} \in \mathcal{F}$, the adversary is expected to find a collision on $H$ subject to $\text{tag}_i = \text{tag}'_i$ after $\mathcal{F}^2$ trials. Therefore if $\mathbf{fa}\left(m, \delta, P_{M_I}^{\max}\right) < \mathcal{F}^2$, this implies that the adversary can find a collision on $H$ with less number of trials. In such a case, the false-accept attack complexity is reduced to the hardness in finding a collision on a standard cryptographic one-way hashing function $H$.

### D. TAG Indistinguishability

Bellare et al., [20] first introduced the notion of key indistinguishability. That is, an adversary has to determine whether the ciphertexts were encrypted by the same key. This notion arose due to the usability concern when the same keys are used across different applications, triggering information leakage. The key privacy notion was later used by Simeons et al. [29] in secure sketch whereby each biometric sketching function is treated as a probabilistic encryption to study the information leakage across multiple sketches [30]. In biometric cryptosystems literature, this notion leads to an attack called attack via record multiplicity (ARM) [10] or correlation attack [31].

In SKE, the stored tag is public. We need to evaluate the advantages gained by an adversary when the same biometric input is used to generate RVs and hence the corresponding tag. Formally, this scenario can be characterized in an *indistinguishability game*. In this game, an adversary holds two distinctive tags and he attempts to learn from the tags to distinguish the tags that are from the same user. The tag that corresponds to a particular biometric should hold this indistinguishability property to ensure no or negligible information can be learned by the adversary.

Given $\mathbf{x} \in \mathbb{R}^d$, nonce $\mathbf{N}$ and $\widehat{\mathbf{N}}$ with a polynomial $f$, the $q$-tag indistinguishability between a challenger and the adversary with parameter $m$ can be portrayed as follows:

1. The challenger generates a biometric vector $\mathbf{x} \in \mathbb{R}^d$ and an RV $\boldsymbol{\phi}_{\mathbf{x}} = \Omega_{\text{IoM}}(\mathbf{x}, m, \mathbf{N})$. Meanwhile, he generates $\widehat{\boldsymbol{\phi}}_{\mathbf{x}} = \Omega_{\text{IoM}}(\mathbf{x}, m, \widehat{\mathbf{N}})$. He then computes the corresponding $\mathbf{TAG}^*$ and sends it to the adversary.
2. The challenger randomly chooses an integer $K \in \{1, \dots, q\}$ and generates a sequence of tags $\{\mathbf{TAG}_1, \dots, \mathbf{TAG}_q\}$. The $K$-th $\mathbf{TAG}$ is generated by $\boldsymbol{\phi}_{\mathbf{x}'} = \Omega_{\text{IoM}}(\mathbf{x}', m, \mathbf{N})$ and $\widehat{\boldsymbol{\phi}}_{\mathbf{x}'} = \Omega_{\text{IoM}}(\mathbf{x}', m, \widehat{\mathbf{N}})$ from another biometric vector $\mathbf{x}' \in \mathbb{R}^d$, where $\text{S}(\mathbf{x}, \mathbf{x}') > \delta$ and $\delta \in (0,1)$ is a fraction of $m$ defined in section IV. The rest of the $q-1$ tags are generated by $\boldsymbol{\phi}_{\mathbf{x}_{R(j)}} = \Omega_{\text{IoM}}(\mathbf{x}_{R(j)}, m, \mathbf{N})$ and $\widehat{\boldsymbol{\phi}}_{\mathbf{x}_{R(j)}} = \Omega_{\text{IoM}}(\mathbf{x}_{R(j)}, m, \widehat{\mathbf{N}})$ from random biometric vectors $\mathbf{x}_{R(j)} \in \mathbb{R}^d$, which hold for $S(\mathbf{x}, \mathbf{x}_{R(j)}) < \delta$ where $j = 1, \dots, q-1$. The challenger then sends $\{\mathbf{TAG}_1, \dots, \mathbf{TAG}_q\}$ to the adversary.
3. The adversary outputs an integer $K' \in \{1, \dots, q\}$ and wins if $K' = K$.

The adversary's advantage in the $q$-tag indistinguishability game can be deduced from the $n$-indistinguishability game discussed in [32]. In our context, we measure $\text{S}(\mathbf{x}, \mathbf{x}')$ and $\text{S}(\mathbf{x}, \mathbf{x}_{R(j)})$ corresponding to the IoM hashing similarity function over $\delta \in (0,1)$. We call the adversary as *SKE-IND adversary* with advantages given as follows:

$$\text{Adv}_{\text{SKE-Ind}} = \frac{q}{q-1}\left|\mathbb{P}[K' = K] - \frac{1}{q}\right|. \quad (2)$$

**Definition 1**: *Under the same IoM hashing (identical random projection matrix set) setup, an SKE is* $(q, \epsilon)$*-indistinguishable in* $\delta \in (0,1)$ *if for any SKE-IND adversary* $\text{Adv}_{\text{SKE-Ind}} \leq \epsilon$ *and perfectly indistinguishable if* $\text{Adv}_{\text{SKE-Ind}} = 0$.

It is crucial to note that the indistinguishability game considers the same secret polynomial $f(.)$ that is protected under SKE through encryption with $\widehat{\boldsymbol{\phi}}_{\mathbf{x}}$. In practice, different secrets might be protected using the same input $\widehat{\boldsymbol{\phi}}_{\mathbf{x}}$ for diverse applications. Nonetheless, as the decoding can succeed only when the number of unique genuine pairs in $\mathbf{U}$ is at least the secret size i.e. $t \geq k$. This implies that as long as $f$ is fixed upon order $k-1$, binding different secrets (coefficient of *f*) has no significant advantage to the adversary. Therefore, our security analysis should hold for both cases. Furthermore, the distribution of the tag or RV highly depends on the biometric vector, and hence it is more appropriate to analyze the information leakage directly from $S(\mathbf{x}, \mathbf{x}')$ with respect to its $\mathbf{TAG}$ or RV. Therefore, the similarity measurement $S(\mathbf{x}, \mathbf{x}')$ becomes an important factor in deriving $\text{Adv}_{\text{SKE-Ind}}$.

While the indistinguishability game consists of sample $\mathbf{x}, \mathbf{x}'$ and $\mathbf{x}_{R(j)}$, it is natural to have different unlocking set sizes $t_{(\mathbf{x},\mathbf{x}')}$, and $t_{(\mathbf{x},\mathbf{x}_{R(j)})}$ corresponding to the different cases when $S(\mathbf{x}, \mathbf{x}') > \delta$, and $S(\mathbf{x}, \mathbf{x}_{R(j)}) < \delta$ for $j = 1, \dots, q-1$, respectively. We define the false-rejection probability $\text{FRP} = 1 - \mathbb{P}(t_{(\mathbf{x},\mathbf{x}')} \geq k)$ and the false-acceptance probability $\text{FAP} = \mathbb{P}\left(t_{(\mathbf{x},\mathbf{x}_{R(j)})} \geq k\right)$ as the measures of information leakage due to $S(\mathbf{x}, \mathbf{x}')$ and $S(\mathbf{x}, \mathbf{x}_{R(j)})$, respectively over the sketches during the $q$-tag indistinguishability game. This information leakage should contribute to the adversary advantages $\text{Adv}_{\text{SKE-Ind}}$.

We now provide a detailed argument on the security of tag-indistinguishability with respect to $\text{Adv}_{\text{SKE-Ind}}$. Suppose that one has nonzero FAP for each secret retrieval attempt. This implies that some minimum amount of information must leak on each secret retrieval attempt. Conversely, the FRP needs to be taken into consideration to avoid overestimation on such information leakage. Suppose an adversary possessing $\mathbf{TAG}^*$ and $\{\mathbf{TAG}_1, \dots, \mathbf{TAG}_q\}$. To distinguish $\mathbf{TAG}^*$ from $\{\mathbf{TAG}_1, \dots, \mathbf{TAG}_q\}$, he needs a biometric vector $\mathbf{x}^*$ that satisfies both $S(\mathbf{x}, \mathbf{x}^*) > \delta$ and $S(\mathbf{x}', \mathbf{x}^*) > \delta$. This eventually allows him to distinguish the $K$-th $\mathbf{TAG}$ from others by simply running a secret retrieval algorithm across all the available tags $\{\mathbf{TAG}_1, \dots, \mathbf{TAG}_q\}$ and $\mathbf{TAG}^*$. Clearly, the adversary can achieve this with minimum effort if $\mathbf{x}^* = \mathbf{x}'$ i.e. by merely focusing on a single case $S(\mathbf{x}, \mathbf{x}^*) > \delta$. If the adversary manages to distinguish the $K$-th sketch with nonnegligible probability i.e. advantage $\text{Adv}_{\text{SKE-Ind}} = \epsilon$, it implies he is able to find $\mathbf{x}^* = \mathbf{x}'$ subject to $S(\mathbf{x}, \mathbf{x}^*) > \delta$ with nonnegligible probability $\epsilon$ as well. Therefore, the notion of tag indistinguishability captures the possibility of a better solution to carry out secret retrieval other than brute-force and false-accept attacks, stimulated by using the $q$-tag indistinguishability game. In such an event, we claim the tag indistinguishability is at most as hard as the false-accept attack, which can be reduced to the hardness in finding a collision on $H$.

**Lemma 1**: *In a q-tag indistinguishability game,* $\text{Adv}_{\text{SKE-Ind}} = \frac{1}{q-1}|\text{FAP} - \text{FRP}|$.

*Proof*: We now argue that the value of $\epsilon$ follows Definition 1. Given that $\delta$ is neither 0 or 1, this means there will be a nonzero FAP for every secret retrieval attempt with $\mathbf{x}^*$ over the $\mathbf{TAG}_{j=1,\dots,q-1}$ generated from $\mathbf{x}_{R(j)}$. Therefore, the *total false-acceptance probability* can be defined as $\text{FAP} = \sum_{j=1}^{q-1} \text{FAP}_j =$ $\sum_{j=1}^{q-1} \mathbb{P}\left(t_{(\mathbf{x}_{R(j)},\mathbf{x})} \geq k\right)$. On the other hand, the secret retrieval would not succeed with probability one but considerable false-rejection probability $\mathrm{FRP} = 1 - \mathbb{P}\left(t_{(\mathbf{x},\mathbf{x}')} \geq k\right)$. Combining these two results leads to $\mathbb{P}[K' = K] = \frac{1}{q}\left(\mathbb{P}\left(t_{(\mathbf{x},\mathbf{x}')} \geq k\right) + \sum_{j=1}^{q-1} \mathbb{P}\left(t_{(\mathbf{x}_{R,j},\mathbf{x})} \geq k\right)\right) = \frac{1}{q}\left(1 - \mathrm{FRP} + \sum_{j=1}^{n-1} \mathrm{FAP}_j - 1\right)$. It follows that $\mathrm{Adv}_{\mathrm{SKE-Ind}} = \frac{1}{q-1}\left|1 - \mathrm{FRP} + \sum_{j=1}^{q-1} \mathrm{FAP}_j - 1\right| = \frac{1}{q-1}\left|\mathrm{FAP} - \mathrm{FRP}\right|$ and the lemma is proved. □

Zero FAR and false-reject rate (FRR) in a biometric cryptosystem with limited query instances do not necessarily imply perfect indistinguishability, because nonzero FAP or FRP may still exist for every matching. Thus, an information leakage measure is required unless one is able to achieve an ideal circumstance such as $\sum_{j=1}^{q-1} \mathbb{P}\left(t_{(\mathbf{x}_{R(j)},\mathbf{x})} \geq k\right) = 0$ and $\mathbb{P}\left(t_{(\mathbf{x},\mathbf{x}')} \geq k\right) = 1$; hence $\mathrm{Adv}_{\mathrm{SKE-Ind}} = 0$. This suggests $S(\mathbf{x}, \mathbf{x}') = 1$ (perfect match) and $S(\mathbf{x}, \mathbf{x}_{R(j)}) = 0$ (complete unlinkability), which would not happen in practice.

We give an illustration to highlight the computation of $\mathrm{Adv}_{\mathrm{IoM-Ind}}$ under certain parameterization.

**Example 3**: Suppose $\mathcal{F} = 251$, $m = 256$, $k = \lceil \delta m \rceil$ with $\delta = \frac{1}{2}$ and $\mathbf{x}'$ is of 80% similar to $\mathbf{x}$ such that $S(\mathbf{x}, \mathbf{x}') = 0.8$. Suppose $\mathbf{x}_{R(j)}$ is only 50% similar with $\mathbf{x}$, so $S(\mathbf{x}, \mathbf{x}_{R(j)}) = 0.5$. Thus $P^2 = (S(\mathbf{x}, \mathbf{x}'))^2$ for the former is $0.8^2$ and the latter is $0.5^2$. It follows $t_{(\mathbf{x},\mathbf{x}_{R(j)})} \sim$ Bin( $m$, $0.5^2$ ) and $t_{(\mathbf{x},\mathbf{x}')} \sim$ Bin($m$, $0.8^2$ ). For $\mathbf{TAG}_{j=1,\ldots,q-1}$ that generated from $\mathbf{x}_{R(j)}$, holds with $S(\mathbf{x}, \mathbf{x}_{R(j)}) = 0.5$, the total false-accept probability is $\mathrm{FAP} = (7.55 \times 10^{-18})(q-1)$. Besides that, $\mathbf{TAG}^*$ that generated from $\mathbf{x}'$ holds $S(\mathbf{x}, \mathbf{x}') = 0.8$, so the $\mathrm{FRP} = 6.30 \times 10^{-5}$. Based on Lemma 1, $\mathrm{Adv}_{\mathrm{SKE-Ind}} = \frac{1}{q-1}\left|(7.55 \times 10^{-18})(q-1) - 1.75 \times 10^{-6}\right|$. Suppose there are $q = 2^{10} \approx 1024$ **TAG** available, $\mathrm{Adv}_{\mathrm{SKE-Ind}} = 1.71 \times 10^{-9}$. These results can further be converted into entropy notion as $-\log(\mathrm{Adv}_{\mathrm{SKE-Ind}}) \sim 30$ bits.

## VI. Experiments & Discussion

In this section, we present a comprehensive performance evaluation on secret retrieval. We adopt fingerprint vectors $\mathbf{x}$ with length 299 generated from the fingerprint minutiae proposed in [24]. The conversion process consists of two sequential stages, i.e. generation of the minutia cylinder-code [33] with the public available MCC SDK, and kernel principal component analysis (KPCA) transformation.

Our experiments were carried out under FVC 2002 [34] (subset DB 1 and 2), FVC 2004 [35] (subset DB 1 and 2), and FVC 2006 [36] (subset DB 1). Each of these datasets has Set A, which contains 100 × 8 fingerprints for FVC2002 and FVC2004 and 140 × 12 fingerprints for FVC2006 and Set B, which contains 10 × 8 fingerprints in FVC2002 and FVC2004 and 10 × 12 fingerprints in FVC2006. We used Set B to generate the KPCA projection matrix and Set A for secret retrieval experiments.

### *A. Experiments for Performance Evaluation*

We used the full FVC protocol for the genuine and imposter attempt test described in [37] as follows:

**Genuine attempt**: For each subject, each fingerprint vector is used for enrollment, and other fingerprint vectors of the same subject are used and try to retrieve the secret with the procedure described in Section III(B). For 100 subjects, the procedure leads to a total number of (100 × 8 × 7)/2=2800 genuine matches, denote as $M_G$ = 2800. For FVC 2006 DB2, we have (140 × 12 × 11)/2=9240 genuine matches.

**Imposter attempt**: For each subject, the first fingerprint vector is used for enrollment, and the second fingerprint vector of other subjects are used to decrypt $\boldsymbol{C}$ for secret retrieval. For 100 subjects, it leads to a total number of (100 × 99)/2=4950 imposter matches, as denoted as $M_I$ = 4950 earlier. For FVC 2006 DB2, we have (140 × 139)/2=9730 imposter matches.

Generally, once $\mathcal{F}$ is fixed, only two major parameters are required to be tuned i.e., $(\delta, m)$. In our exposition, we simply fixed $\mathcal{F} = 251$ for the experiments. Because both $\mathcal{F}$ and $m$ are the parameters associated to IoM hashing, we fixed $m = 1024$ as it is the best parameter for $\mathcal{F} = 251$ under FVC datasets reported in [19]. We repeated the experiments with different $\delta = \frac{k}{m}$ with $k$ = 8, 12, …, 24. The FRR and FAR of the experiment results are tabulated in Table I.

From Table I, we observe similar performance trajectories for FAR and FRR for all datasets when $k$ (or $\delta$) grows large. Larger $k$ increase the FRR. On the other hand, small $k$ does improve FRR but unfortunately, impairs FAR. Nevertheless, FAR remains extremely small for FVC 2002 and 2004 datasets regardless $k$, i.e. mostly are zero due to the strong resilience property of SKE justified in Section IV. Only for FVC 2006 DB2, one expects to have $k > 24$ to attain a lower FAR.

TABLE I: Experiment Results For SKE For Various *k*

| **FVC 2002 DB1** | $k$ =8 | $k$ =12 | $k$ =14 | $k$=16 | $k$=18 | $k$=20 | $k$=24 |
|---|---|---|---|---|---|---|---|
| FRR(%) | 0.29 | 0.38 | 0.38 | 0.52 | 0.61 | 0.67 | 0.76 |
| FAR(%) | 0.14 | 0.02 | 0.00 | 0.00 | 0.00 | 0.00 | 0.00 |
| EER(%) | 0.75 | 0.75 | 0.75 | 0.75 | 0.75 | 0.75 | 0.75 |
| **FVC 2002 DB2** | $k$ =8 | $k$ =12 | $k$ =14 | $k$=16 | $k$=18 | $k$=20 | $k$=24 |
| FRR(%) | 0.33 | 0.33 | 0.38 | 0.48 | 0.48 | 0.48 | 0.62 |
| FAR(%) | 0.14 | 0.00 | 0.00 | 0.00 | 0.00 | 0.00 | 0.00 |
| EER(%) | 0.30 | 0.30 | 0.30 | 0.30 | 0.30 | 0.30 | 0.30 |
| **FVC 2004 DB1** | $k$ =8 | $k$ =12 | $k$ =14 | $k$=16 | $k$=18 | $k$=20 | $k$=24 |
| FRR(%) | 5.90 | 9.10 | 10.66 | 11.89 | 12.71 | 13.81 | 15.43 |
| FAR(%) | 3.95 | 0.95 | 0.52 | 0.38 | 0.23 | 0.14 | 0.09 |
| EER(%) | 4.46 | 4.46 | 4.46 | 4.46 | 4.46 | 4.46 | 4.46 |
| **FVC 2004 DB2** | $k$ =8 | $k$ =12 | $k$ =14 | $k$=16 | $k$=18 | $k$=20 | $k$=24 |
| FRR(%) | 8.33 | 10.24 | 11.19 | 11.91 | 12.57 | 13.76 | 15.05 |
| FAR(%) | 2.72 | 0.89 | 0.50 | 0.42 | 0.28 | 0.22 | 0.12 |
| EER(%) | 6.64 | 6.64 | 6.64 | 6.64 | 6.64 | 6.64 | 6.64 |
| **FVC 2006 DB2** | $k$ =8 | $k$ =12 | $k$ =14 | $k$=16 | $k$=18 | $k$=20 | $k$=24 |
| FRR(%) | 0.73 | 1.22 | 1.52 | 1.75 | 2.06 | 2.38 | 3.06 |
| FAR(%) | 1.24 | 0.39 | 0.26 | 0.16 | 0.14 | 0.11 | 0.06 |
| EER(%) | 0.86 | 0.86 | 0.86 | 0.86 | 0.86 | 0.86 | 0.86 |

TABLE II: VARIOUS BENCHMARKINGS FOR SKE

| Dataset | Equal Error Rate (EER, %) (error rate where FAR = FRR) | | |
|---|---|---|---|
| | Fingerprint Vectors [24] (Cosine similarity measure) | IoM hashed vector ($m = 1024, \mathcal{F} = 256$) (Normalized Hamming distance over finite field) | SKE ($m = 1024, \delta = \frac{8}{1024}, \mathcal{F} = 251$) |
| FVC2002 DB1 | 0.56 | 0.75 | 0.75 |
| FVC2002 DB2 | 0.28 | 0.28 | 0.30 |
| FVC2004 DB1 | 4.29 | 4.73 | 4.46 |
| FVC2004 DB2 | 6.98 | 6.82 | 6.64 |
| FVC2006 DB2 | 0.70 | 0.72 | 0.86 |

TABLE III: AVERAGE ENCODING AND DECODING RATES (S) OF SKE PARAMETERS SET UP $m = 1024,\ \mathcal{F} = 251$ FOR VARIOUS $k$

| | Average Encoding Time (sec) | | | | Average Genuine/ Imposter Decoding Time (sec) | | | |
|---|---|---|---|---|---|---|---|---|
| Dataset | $k = 8$ | $k = 18$ | $k = 36$ | $k = 54$ | $k = 8$ | $k = 18$ | $k = 36$ | $k = 54$ |
| FVC2002 DB1 | 0.425 | 0.954 | 1.612 | 2.645 | 0.361/ 0.185 | 0.453/ 0.160 | 0.678/ 0.178 | 0.985/ 0.183 |
| FVC2002 DB2 | 0.471 | 0.815 | 1.710 | 2.412 | 0.593/ 0.178 | 0.6115/ 0.173 | 0.737/ 0.170 | 1.004/ 0.177 |
| FVC2004 DB1 | 0.434 | 0.858 | 1.560 | 2.256 | 0.354/ 0.177 | 0.389/ 0.165 | 0.469/ 0.163 | 0.652/ 0.208 |
| FVC2004 DB2 | 0.422 | 0.916 | 1.584 | 2.255 | 0.401/ 0.187 | 0.387/ 0.169 | 0.454/ 0.167 | 0.595/ 0.165 |
| FVC2006 DB2 | 0.444 | 0.903 | 1.677 | 2.554 | 0.533/ 0.213 | 0.641/ 0.310 | 0.804/ 0.238 | 0.939/ 0.171 |

### *B. Performance Preservation of SKE*

In this section, we demonstrate that SKE with its best parameters manages to preserve the accuracy performance as in the original fingerprint vector. Note this is not a trivial task as the performance of the original biometric system is often not preserved, as the errors will be accumulated when propagated to the encryption domain of biometric cryptosystems. Table II illustrates the relative performances of three systems in different domains i.e. biometric vectors (feature domain), IoM hashed vectors (max ranked domain), and SKE (encryption domain) in terms of equal error rate (EER). Under the same evaluation protocol for genuine and imposter attempts, to our surprise, we noticed that SKE attains better performances under FVC2002 DB1, FVC2004 DB1, FVC2004 DB2, and FVC2002 DB2 while FVC2006 DB2 degraded slightly. The performance gain could be attributed to the use of the long IoM hashed vector ($m$=1024), error checking mechanism with **TAG**, and Shamir's secret-sharing scheme.

### *C. Computation time of SKE*

Table III illustrates the average encoding time defined as $\frac{\text{total time taken}}{\text{No. enrollment}}$ during enrollment and average decoding time defined as $\frac{\text{total time taken}}{\text{No. retrieval trials}}$ according to the genuine and imposter attempts with respect to different $k$. It is noted that the average encoding and decoding times for genuine attempt increase when $k$ is tuned larger. On the other hand, the decoding time for imposter attempts is close to mean $0.187 \pm 0.12$ s regardless of $k$ value. This observation supports the claim that one would expect that only a genuine user can retrieve the secret with a large amount of unique genuine pairs.

### *D. Achievable Security over Specific Parameter*

In an ideal circumstance, the SKE can achieve formal security favorably as illustrated in Section V. However, due to the system performance–security trade-off, it would be good to benchmark the achievable security based on the FVC datasets with respect to their best parameters (i.e. lowest FRR at FAR = 0%).

Recall that the brute-force and false-accept attack complexity estimation required the knowledge of maximum collision probability $P_{M_I}^{\max}$ over $M_I$ imposter matches between different RVs. We therefore have recorded the values of $P_{M_I}^{\max}$ obtained for different FVC fingerprint datasets with $M_I$ = 4950, which are $P_{M_I}^{\max}$ =0.1006, 0.1279, 0.2047, 0.1846 for FVC2002 DB1, DB2, and FVC2004 DB1, DB2 respectively. We opt $m = 1024$, $k = 54$ ($\delta = \frac{54}{1024}$), and $\mathcal{F} = 251$, which are exactly the parameters used in Table IV and V, for the following security evaluations:

**Brute-Force Attack (Section VA)**: BF can be estimated with $-\log(\mathcal{F}^k) = -\log(251^k) \geq 256$ bits.

**False-Accept Attack (Section VB):** The maximum false-accept attack complexity (or maximum achievable system entropy) can be estimated from $\mathbf{fa}(m, \delta, P_{M_I}^{\max}) = -\log\left(\mathbb{P}\left(t_{(P_{M_I}^{\max})} > \lceil \delta m \rceil\right)\right)$ as 71 bits, 43 bits, 10 bits, and 5 bits for FVC2002 DB1, DB2, and FVC2004 DB1, DB2, respectively. Observe that the false-accept attack complexity is surprisingly low for a noisy dataset such as FVC2004 DB1, DB2.

**TAG Indistinguishability (Section VC)**: Follow Lemma 1, the advantages for *SKE-IND adversary* can be estimated from $\text{Adv}_{\text{SKE-Ind}} = \frac{1}{q-1}|\text{FAP} - \text{FRP}|$ where $\text{FAP} = \sum_{j=1}^{q-1} \text{FAP}_j = \sum_{j=1}^{q-1} \mathbb{P}\left(t_{(\mathbf{x}_{R(j)},\mathbf{x})} \geq k\right)$ and $\text{FRP} = 1 - \mathbb{P}\left(t_{(\mathbf{x},\mathbf{x}')} \geq k\right)$. Because $t_{(\mathbf{x},\mathbf{x}_{R(j)})} \sim \text{Bin}(m, \text{S}(\mathbf{x},\mathbf{x}_{R(j)})^2)$ and $t_{(\mathbf{x},\mathbf{x}')} \sim \text{Bin}(m, \text{S}(\mathbf{x},\mathbf{x}')^2)$, compute $t_{(\mathbf{x},\mathbf{x}_{R(j)})}$ and $t_{(\mathbf{x},\mathbf{x}')}$ require precise knowledge on $\text{S}(\mathbf{x},\mathbf{x}_{R(j)})$ and $\text{S}(\mathbf{x},\mathbf{x}')$. Accordingly, it is natural to adopt $P_{M_I}^{\max} = \text{S}(\mathbf{x},\mathbf{x}_{R(j)})$ for $j = 1, \dots, q-1$. On the contrary, $\bar{P}_{M_G} = \text{S}(\mathbf{x},\mathbf{x}')$ is opted to estimate $\mathbb{P}\left(t_{(\mathbf{x},\mathbf{x}')} \geq k\right)$, and so $\text{FRP} = 1 - \bar{P}_{M_G}$, where $\bar{P}_{M_G}$ is the average genuine matching scores of the RV pairs over $M_G = 2800$ of different query biometric vectors $\mathbf{x}'$. We take FVC 2002 DB1 as an instance of estimation where $P_{M_I}^{\max} = 0.1006$ and $\bar{P}_{M_G} = 0.5944$. Given $q = 2^{10}$, $\text{FAP} = (2^{10} - 1)(4.53 \times 10^{-22})$ and $\text{FRP} = 1.11 \times 10^{-16}$ can be computed, hence $\text{Adv}_{\text{SKE-Ind}} = 1.081 \times 10^{-19}$ or its entropy form $-\log(\text{Adv}_{\text{SKE-Ind}}) = 64$ bits, which is lower than the false-accept complexity at 71 bits.

Apart from this, it is also numerically verifiable that increment of $q$ would eventually lead to lower $\text{Adv}_{\text{SKE-Ind}}$ or higher $-\log(\text{Adv}_{\text{SKE-Ind}})$. For instance, with $q = 2^6, 2^{10}, 2^{20}, 2^{30}, 2^{40}, 2^{50}$, the $-\log(\text{Adv}_{\text{SKE-Ind}})$ = 59, 64, 72, 71, 71, 71 bits, respectively and sooner will be bounded by the false-accept attack complexity. This suggests the presence of a more efficient algorithm with computation complexity lower bounded to false-accept attack complexity. Noticeably, this statement holds for any two distinct $\mathbf{TAGs}$ sourced from similar biometric vectors over $q$-number of $\mathbf{TAGs}$. Lower $q$ indicates that the adversary has a higher chance in retrieving the secret.

TABLE IV: COMPARISON WITH EXISTING SCHEMES UNDER FULL FVC PROTOCOL WITH PARAMETER SET $m = 1024, \mathcal{F} = 251$, AND SPECIFIED $k = 12, 31, 36$, AND $45$.

| | FVC2002 | | FVC2004 | | FVC2006 |
|---|---|---|---|---|---|
| | DB1 | DB2 | DB1 | DB2 | DB2 |
| | FRR (%) (FAR=0) /EER (%) | FRR (%) (FAR=0) /EER (%) | FRR (%) (FAR=0) /EER (%) | FRR (%) (FAR=0) /EER (%) | FRR (%) (FAR=0) /EER (%) |
| Yang et al. [44] | ≥35.8/11.8 | >26.8/10.4 | -/- | ≥57.6/20.6 | -/- |
| Yang et al. [39] | -/4.5 | -/5.99 | -/- | ≥ 26.1/- | -/3.07 |
| Li et al. [38] | 16.6/0 | 11.5/0 | -/- | ≥24.8/- | - |
| Li et al. [9] | /31.2 | -/27.7 | -/- | -/- | -/- |
| **SKE ( $k$=12)** | **0.38/0.75** | **0.38/0.30** | **-/4.56** | **-/6.64** | **-/0.86** |
| **SKE ($k$=31)** | **0.92/0.75** | **0.84/0.30** | **18.48/4.56** | **-/6.64** | **-/0.86** |
| **SKE ($k$=36)** | **0.92/0.75** | **0.99/0.30** | **22.7/4.56** | **18.04/6.64** | **-/0.86** |
| **SKE ($k$=45)** | **0.95/0.75** | **1.23/0.30** | **24.3/4.56** | **20.14/6.64** | **5.46/0.86** |

TABLE V: COMPARISON WITH EXISTING SCHEMES UNDER 1VS1 PROTOCOL WITH PARAMETER SET $m = 1024, \mathcal{F} = 251$, AND SPECIFIED $k = 12, 18$, AND $54$.

| | FVC2002 | | FVC2004 | | FVC2006 |
|---|---|---|---|---|---|
| | DB1 | DB2 | DB1 | DB2 | DB2 |
| | FRR (%) (FAR=0) /EER (%) | FRR (%) (FAR=0) /EER (%) | FRR (%) (FAR=0) /EER (%) | FRR (%) (FAR=0) /EER (%) | FRR (%) (FAR=0) /EER (%) |
| Nandakumar et al. [11] | -/- | 14.0 | -/- | -/- | -/- |
| Li et al. [9] | 14.0/- | 8.0/- | -/- | -/- | -/- |
| Yang et al. [44] | ≥8.0/3.38 | >6.0/0.59 | -/- | ≥41/14.88 | -/- |
| Yang et al. [39] | 4.0/0.0 | 2.0/1.02 | -/- | ≥24.72/- | -/4.83 |
| Yang et al. [41] | 15.0/0.0 | 7.0/0.0 | -/- | -/- | |
| Tams et al. [13] | -/- | 21.0/0.0 | -/- | -/- | -/- |
| Li et al. [38] | 2.0/0.0 | 0.0/0.0 | -/- | 23.4/- | -/- |
| Nagar et al [42] | - | 7.00/0.0 | -/- | -/- | -/- |
| Nagar et al. [43] | - | -/1.00 | -/- | -/- | -/- |
| **SKE ($k$=12)** | **3.0/0.81** | **-/0.00** | **-/3.01** | **-/3.94** | **-/1.06** |
| **SKE ($k$=18)** | **4.0/0.81** | **0.00/0.00** | **-/3.01** | **-/3.94** | **-/1.06** |
| **SKE ($k$=54)** | **4.0/0.81** | **2.00/0.00** | **20.0/3.01** | **17.0/3.94** | **8.57/1.06** |

*Remarks*: Recall that we have relaxed our security requirement over a smaller imposter match size ($M_I$), which is crucial for $P_{M_I}^{\max}$ estimation. The achievable security does not necessary reflect the actual achievable security bound but merely for better resilience. In other words, the achievable security can be improved through the increment of three parameters $\mathcal{F}, m,$ and $\delta(\text{or } k)$ ) but trade with an accuracy performance degradation. Especially, the field size $\mathcal{F}$ that links to the secret size is also an important concern in finding a collision with any cryptographic one-way hash function $H$. Nevertheless, such a relaxation in the security is necessary to show more convincing evaluation, especially when the biometric samples are limited due to the difficulty in data collection.

### *E. Comparison with Stat-of-the-art*

Given that SKE shares similar ingredients with fuzzy vault where both use a polynomial reconstruction mechanism, we provide a numerical comparison with state-of-the-art fuzzy vault implementations. Table IV tabulates the comparison result based on the full FVC testing protocol for genuine and imposter attempts. Comparison results in Table V are based on the 1vs1 protocol for genuine attempts (e.g. the first sample is used to enroll while the second sample is used for secret retrieval) and the full FVC protocol for imposter attempts.

From Table V, we choose SKE with $k$ = 12, 18, and 54 to compare with the competing techniques. The former outperforms the rest with higher system security (we used residue entropy in our case) while the latter shows higher FRR compare to the works proposed by Li et al. [38] and Yang et al. [39] under FVC2002 DB1 and DB2.

By following the full FVC protocol, which is considered more difficult than the 1vs1 protocol, SKE with $k$ = 12, 31, 36, and 45 outperforms others with lower FRRs at FAR = 0% (Table IV).

## VII. TEMPLATE PROTECTION PROPERTIES OF SKE

SKE as a biometric cryptosystem instance can also be perceived as a BTP scheme [22] [23]. The BTP must satisfy four criteria, namely non-invertibility, unlinkability, revocability, and performance.

In this section, we evaluate SKE on each of these properties, mostly focusing on the RV of RSV (e.g. IoM hashing).

### *A. Noninvertibility*

A BTP scheme should be computationally infeasible to reverse engineer the original biometric data from its protected template or/and the helper data i.e. sketch and TAG in SKE. This prevents the privacy invasion and security attacks on the biometric systems. This property is well covered by the TAG indistinguishability in Section VC. To be precise, we measure the adversary advantage for an adversary trying to compromise the SKE through any potential strategies not limited to brute force and false accept. We find out given an adversary with pre-knowledge that two different TAGs among $q$ number of tags must source from the same user, he/she eventually gains advantages and is able to reconstruct the secret with complexity lower bounded to the false-accept attack. As we shall see, our results show that with $q = 2^6, 2^{10}, 2^{20}, 2^{30}, 2^{40}, 2^{50}$, the computed adversary advantages (in bits unit) are 59, 64, 72, 71, 71, 71 respectively for FVC 2002 DB1. This complexity directly implies a compromise of the input biometric data without the necessity to reverse engineer the input.

## *B. Revocability*

The revocability enables a protected template to be revoked once compromised and to be replaced with a new one. To evaluate the revocability property of SKE empirically, FVC 2002 DB1 was used for evaluation. Note that **TAG** in SKE is merely constructed by RV (or IoM hashed vector), and therefore IoM hashing is used in the experiment. The first fingerprint sample of each subject was used to generate a total number of eight distinctive RVs with different random Gaussian matrices (e.g. eight numbers of different nonce $\mathbf{N}_1, \ldots, \mathbf{N}_8$ ). Each RV was used for SKE enrollment. During retrieval, we followed the same protocol as discussed in Section VIA for imposter and genuine attempts. A total number of 4950 trials for imposter attempts and 2800 trials for genuine attempts were recorded. However, because different nonce is used, we named the genuine attempt (under same subject) as *pseudo imposter attempt* follows [19].

Figure 3 shows the pseudo imposter score and imposter score, which corresponds to the normalized genuine pairs count. As expected, the distribution of both pseudo imposter and imposter attempts are highly similar and overlapped. This suggests that the RV derived from the newly generated RVs with different nonce over the same subject are indistinguishable from the RV generated from others. Therefore, the new RV generation indeed can be used to replace the old one and revocability is satisfied.

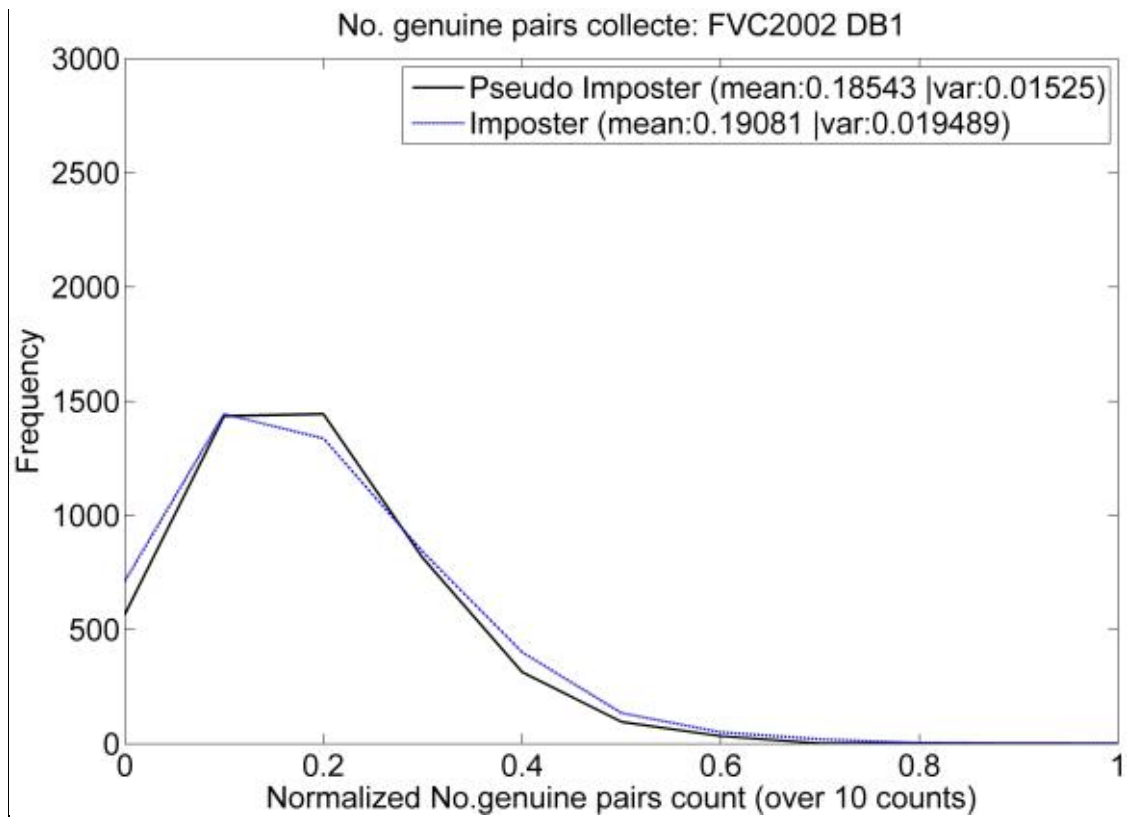

Fig. 3: Pseudo imposter score and imposter score in FVC2002 DB1

## *C. Unlinkability*

The unlinkability criterion demands that the protected biometric data should not be differentiated whether they are generated from the same user's biometrics. This is to prevent matching across different applications (cross-matching). For unlinkability evaluation of SKE, we followed the newly proposed unlinkability framework by Gomez-Barrero et al., (2018) [40].

The unlinkability evaluation was carried out with FVC2002 DB1, FVC2002 DB2, FVC2004 DB1 and FVC2004 DB2. To do so, for every sample in the dataset, different nonce were applied to generate the RV pair $(\boldsymbol{\phi}_\mathbf{x}, \widehat{\boldsymbol{\phi}}_\mathbf{x})$. Each RV pair was used for SKE enrollment.

Let $\mathbb{P}(s|H_m)$ and $\mathbb{P}(s|H_{nm})$ be the probability densities of a given similarity score $s \in [0,1]$ that being computed from mated samples, $H_m$ i.e. same subjects and nonmated samples $H_{nm}$ i.e. different users, respectively. In our exposition, $\mathbb{P}(s|H_m)$ and $\mathbb{P}(s|H_{nm})$ refer to the probability density of the normalized number of revealed genuine pairs (equivalent to similarity score $s$ ) for genuine and imposter attempts, respectively. A total number of 2800 genuine attempts and 4950 imposter attempts for each dataset were recorded.

Given a likelihood ratio $LR(s) = \mathbb{P}(s|H_m)/\mathbb{P}(s|H_{nm})$, the unlinkability property can be characterized by the local linkability measure $\mathrm{D}_{\leftrightarrow}(s)$ and the global linkability measure $\mathrm{D}_{\leftrightarrow}^{sys}$ defined as:

$$\mathrm{D}_{\leftrightarrow}(s) = 2\,\frac{LR(s)\cdot\omega}{1+LR(s)\cdot\omega} - 1, \tag{3}$$

where we set $\omega \in [0,1] = 1$ and

$$\mathrm{D}_{\leftrightarrow}^{sys} = \int \mathbb{P}(s|H_m)\cdot\,\mathrm{D}_{\leftrightarrow}(s)ds\,. \tag{4}$$

In brief, $\mathrm{D}_{\leftrightarrow}(s) \in [0,1]$ reports an increasing degree of linkability. Thus, high $\mathrm{D}_{\leftrightarrow}(s)$ suggests the revelation of the genuine pair through authenticating **TAG**(**TAG'**) within mated samples. Figure 4 depicts $\mathbb{P}(s|H_m)$ and $\mathbb{P}(s|H_{nm})$ under genuine attempt (mated) and imposter attempt (nonmated) for secret retrieval on the four datasets. It is clearly shown that the score distribution for mated and nonmated instances are highly overlapped. This implies that it is unable for an adversary to differentiate whether the RV pairs $(\boldsymbol{\phi}_{\mathbf{x}'}, \widehat{\boldsymbol{\phi}}_{\mathbf{x}'})$ and $(\boldsymbol{\phi}_\mathbf{x}, \widehat{\boldsymbol{\phi}}_\mathbf{x})$ (sourced from different biometric vector $\mathbf{x}$ and $\mathbf{x}'$ respectively) are generated from the same fingerprint samples or not. Additionally, the near to zero values of $\mathrm{D}_{\leftrightarrow}^{sys}$ for four datasets i.e. $0.0406, 0.0290, 0.0692,$ and $0.0126$ of FVC2002 DB1, FVC2002 DB2, FVC2004 DB1 and FVC2004 DB2, respectively suggest that the linkability of the RVs pairs used is highly unlikely.

## *D. Performance*

The performance criterion of BTP demands that the accuracy performance of the protected system should not be poorer than its original counterpart. The performance preservation from fingerprint vector to IoM hashed vector and to SKE has been well discussed under Section VI(B) and the results are tabulated in Table II.

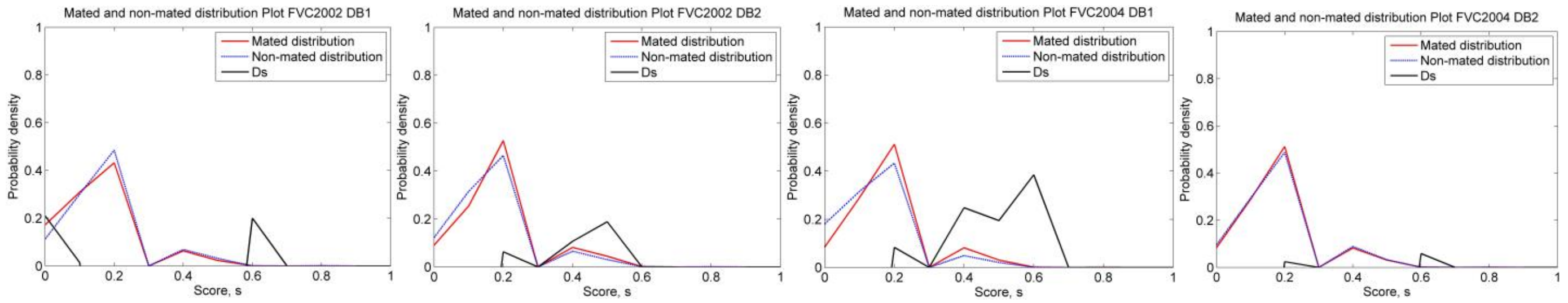

Fig. 4: Mated, nonmated score distributions and $\mathrm{D}_{\leftrightarrow}(s)$ plots for FVC2002 DB1, DB2, FVC2004 DB1 and B2 (left to right).

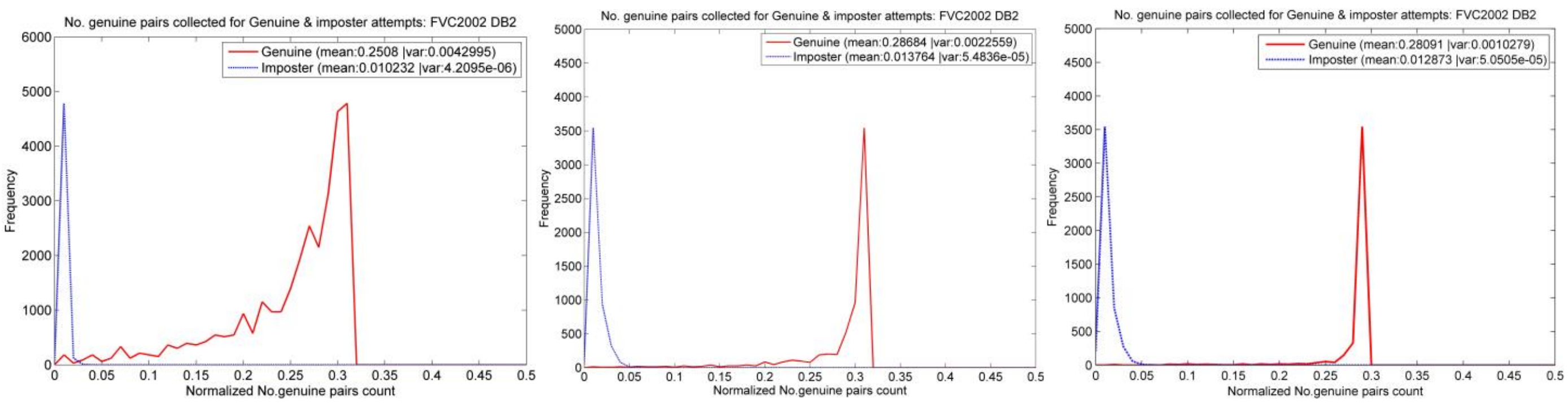

(a) (b) (c)

Fig. 5: Normalized revealed genuine pairs count for genuine and imposter attempts in FVC2002 DB2: (a) $m = 256$; (b) $m = 512$; (c) $m = 1024$.

Besides, we also provide a discussion on the parameter $m$, which is another critical factor to determine the SKE performance. Note that the number of revealed unique genuine pairs in $\mathbf{U} = \{(\varphi_{\mathrm{x}(j)}, f(\varphi_{\mathrm{x}(j)}))\}_{j=1}^{t}$ follows $t \sim \mathrm{Bin}(m, P^2)$ and $\mathbb{E}(t) = mP^2$ (Section IV). By this means, the effect of $m$ can be empirically examined through observing the mean values of the number of revealed unique genuine pairs for genuine and imposter attempts. Figure 5 shows that the means of both genuine (red) and imposter (blue) curves *remain unchanged* irrespective to $m$ whereas the variances *shrink when* $m$ *increases* from 128, 512, and 1024. Hence, the net effect is the gradual separation of genuine and imposter curves when $m$ becomes large, resulting in a better secret retrieval rate in terms of FAR and FRR.

## VIII. Conclusion

In this paper, we proposed an SKE scheme for vectorial biometric secret binding. The SKE is composed of an RV pair that resembles the encryption-decryption key pair in symmetric cryptosystems, along with a simple filtering mechanism and Shamir's secret-sharing scheme for error checking and correction. The SKE possesses a strong resilient property and was empirically validated by the five subsets of FVC fingerprint benchmark. However, the scheme is not limited to fingerprint and can be applied to any biometric features that is manifested in the vector form. The security threat of SKE was also established and analyzed upon the random error model, which can be characterized by the similarity measure over the input biometric vector by resilient set vectorizer (IoM hashing). In a nutshell, we offer an alternative to exploit the biometric input structure while RSV, which implicitly allows one to measure the similarity of the input biometric vectors. Such an exploitation is crucial, especially for more complete security evaluation over the non-uniformity of the input biometric. We believe that our security model will benefit future cryptographic research on biometrics, whereby error tolerance and nonuniformity must be taken into consideration.